\pdfoutput=1
\documentclass[11pt]{article}
\usepackage{acl}
\usepackage{times}
\usepackage{latexsym}
\usepackage[T1]{fontenc}
\usepackage[utf8]{inputenc}
\usepackage{microtype}
\usepackage{inconsolata}
\usepackage{graphicx}
\usepackage{amsmath}
\usepackage{dirtytalk}
\usepackage{booktabs}
 \usepackage{subfigure}
\usepackage{placeins}
\usepackage{subcaption}
\usepackage{CJKutf8}
\usepackage{float}
\usepackage{hyperref}

\DeclareFontFamily{\encodingdefault}{\ttdefault}{\hyphenchar\font=`\-}

\title{Different types of syntactic agreement recruit the same units \\ within large language models}

\author{
  Daria Kryvosheieva$^{1}$,
  Andrea de Varda$^{1}$,
  Evelina Fedorenko$^{1}$,
  Greta Tuckute$^{1,2}$ \\
  $^1$Massachusetts Institute of Technology \\
  $^2$Kempner Institute at Harvard University \\
  \texttt{\{daria\_k, devar\_ag, evelina9\}@mit.edu}, 
  \texttt{gtuckute@fas.harvard.edu}
}

\begin{document}
\defcitealias{tii-2024}{TII (2024)}
\maketitle
\begin{abstract}
Large language models (LLMs) can reliably distinguish grammatical from ungrammatical sentences, but how grammatical knowledge is represented within the models remains an open question. We investigate whether different syntactic phenomena recruit shared or distinct components in LLMs. Using a functional localization approach inspired by cognitive neuroscience, we identify the LLM units most responsive to 67 English syntactic phenomena in seven open-weight models. These units are consistently recruited across sentences containing the phenomena and causally support the models' syntactic performance. Critically, different types of syntactic agreement (e.g., subject-verb, anaphor, determiner-noun) recruit overlapping sets of units, suggesting that agreement constitutes a meaningful functional category for LLMs. This pattern holds in English, Russian, and Chinese; and further, in a cross-lingual analysis of 57 diverse languages, structurally more similar languages share more units for subject-verb agreement. 
Taken together, these findings reveal that syntactic agreement---a critical marker of syntactic dependencies---constitutes a meaningful category within LLMs' representational spaces.\footnote{Code and data available via \href{https://github.com/dariakryvosheieva/syntax-units}{github.com/dariakryvoshe\-ieva/syntax-units}}

\end{abstract}

\section{Introduction}

Large language models (LLMs) can now generate well-formed sentences that are often indistinguishable from those produced by humans \citep{jones2024does,jones2025large}. Ample evidence suggests that this generative capacity is supported by syntactic generalization---the models' ability to apply grammatical constraints in new contexts (e.g., \citealp{gender, futrell2019neural, gulordava-2018, wilcox2018rnn, BERTNPI}; though see \citealp{marvin2018targeted, mccoy2020right} for evidence of systematic limitations). However, it remains unclear how grammatical rules are represented and implemented within the models. For example, does applying a particular rule consistently engage the same subset of the model? Do different rules recruit the same model units? In other words, is there a subset of the model that performs syntactic computations across grammatical phenomena, or are different rules instead instantiated in distinct parts of the model? And is syntactic processing implemented in the same units across different languages?

In NLP, such \say{how} questions are increasingly tackled through mechanistic interpretability approaches \citep{mueller2024quest}. Two main approaches dominate this line of work \citep{rai2024practical, wang2025lost}: one focuses on \textit{features}, i.e., identifying human-interpretable properties encoded in model representations; the other focuses on \textit{circuits}, i.e., tracing the paths and components in the computation graph that implement specific behaviors. In this paper, we adopt the feature-based approach, and in particular, we draw inspiration from a common strategy in cognitive neuroscience: functional localization. This approach \citep{saxe2006divide} aims to isolate regions of the brain that support particular cognitive capacities: for example, to identify brain regions that process faces, a contrast between faces and non-face objects may be used, and to identify brain regions that process words and syntactic structure, a contrast between sentences and lists of non-words may be used \citep{fedorenko2010new}.
Recent work has shown that functional localization can be applied to LLMs to identify specialized sets of units: \citet{alkhamissi-2025} adapted standard neuroimaging `localizer’ contrasts, like those above, to LLMs, identifying the top-$k$ units that best distinguish between language (sentences) and control stimuli (non-word lists). They then showed that ablating these units impairs language performance more than ablating the same number of non-language-selective units, demonstrating parallels to lesion studies in humans. Here, we apply a similar approach to a narrower domain of language processing—the processing of syntactic structure—to illuminate the architecture underlying LLMs’ grammatical competence.

Our study makes the following main contributions:
\begin{itemize}
\item \textbf{Syntax-responsive units.} We identify units in seven open-weight LLMs that are engaged in each of 67 syntactic phenomena, generalize across sentences containing that phenomenon, and are causally implicated in model behavior for the phenomenon.
\item \textbf{Agreement as a functional category.} Unlike some syntactic phenomena, different types of syntactic agreement in English (subject-verb, anaphor, and determiner-noun agreement) recruit overlapping units, suggesting that these phenomena draw on shared resources within LLMs.
\item \textbf{Cross-linguistic agreement overlap.} The pattern replicates in Russian and Chinese: different types of agreement again draw on overlapping units. In addition, agreement-responsive units show partial overlap \textit{across} languages, and in an analysis of 57 languages, the extent of this cross-lingual overlap increases with syntactic similarity---more similar languages share more units for agreement.
\end{itemize}

\section{Related work}

Previous work has shown that LLMs are sensitive to diverse aspects of syntactic structure, such as number and gender agreement, verb argument structure, and syntactic dependency structure, assigning lower probabilities to ungrammatical sentence variants \citep{gulordava-2018, gender, wilcox2018rnn, BERTNPI,linzen-2016, gauthier-2020}. These studies treat LLMs as \say{behavioral} participants, probing their syntactic expectations through surprisal in controlled minimally different sentence pairs. This approach has revealed interesting generalizations, but it leaves open the question of \textit{how} grammatical ability is instantiated within a model. Addressing this question requires examination of the model’s internal representations.


Model interpretability work has investigated how syntactic information is represented within LLMs. Some early studies showed that syntax-related information is encoded in the intermediate layers of language models, as revealed by classifiers trained to recover syntactic labels from hidden states \citep{belinkov2017evaluating, tenney-etal-2019-bert, jawahar-etal-2019-bert}. Other studies have examined the geometry of embedding spaces, showing that syntactic dependency structure can be recovered from representational distances. In particular, \citet{hewitt-manning-2019-structural} introduced an approach (`structural probe’) that reveals that entire parse trees are implicitly encoded in model representations. Also leveraging the geometry of the embedding space, \citet{starace-etal-2023-probing} found that part-of-speech and dependency information are jointly represented.

More recent work has moved from distributed representations to identifying sparse, functionally meaningful components. \citet{lakretz2019emergence, lakretz2021mechanisms} used targeted ablations in LSTMs to identify model units that are responsible for number agreement. Complementing these findings, \citet{mueller-etal-2022-causal} applied counterfactual interventions to neuron activations to study how models encode agreement dependencies, \citet{boguraev2025causal} used causal interventions to characterize shared mechanisms across filler-gap constructions, and \citet{lu-etal-2025-paths} used interventions to improve factual recall accuracy in non-English languages. \citet{marks2024sparse} and \citet{leporiuncovering} isolated minimal subgraphs within models---so-called sparse feature circuits---that support subject-verb agreement. Using a feature-based approach, \citet{durrani2020analyzing} identified subsets of units predictive of performance on various syntactic tasks, such as part-of-speech tagging and chunking a text into constituents. 
\citet{duan2025syntax} introduced a syntactic selectivity index (SSI), which quantifies how strongly a model unit differentiates grammatical versus ungrammatical sentences for a given syntactic phenomenon relative to other phenomena based on activation magnitude. They then tracked how these high-SSI units emerge over the course of training, and showed that ablating high-SSI units increases perplexity on grammatical sentences.

Prior work on multilingual models has shown that different languages rely on shared representations of syntactic structure. At the level of output probabilities, \citet{michaelov2023structural} and \citet{arnett2025acquisition} have studied cross-lingual generalization using structural priming approaches.
At the level of full model representations, (morpho)syntactic features are encoded in shared geometries and patterns of unit activations \citep{chi-etal-2020-finding,papadimitriou2021deep,brinkmann2025large}, and at a more granular level, syntax-sensitive neurons and circuits generalize across languages (\citealp{devarda2024syntax, stanczak-etal-2022-neurons, ferrando2024similarity, zhang2025the}; cf. \citealp{tang2024language}).

Together, these studies suggest that syntactic information is not only recoverable from LLM representations, but often implemented via functionally-specific components, such as subnetworks or even units, for some phenomena. However, most prior work has either targeted specific constructions---often subject-verb agreement (cf. \citealp{duan2025syntax})---or has lacked a unifying localization framework across experimental manipulations. Thus, past work leaves open a critical question: do syntax-responsive model components encode general syntactic knowledge, relevant to diverse grammatical phenomena, or are they narrowly tuned to specific syntactic relations/rules? To answer this question, we systematically identify the most relevant model units for each of dozens of syntactic phenomena (i.e., units that best differentiate grammatical and ungrammatical variants for the phenomenon in question), test these units' causal role in grammatical ability via ablation, and examine overlap among the units that are most relevant to different grammatical phenomena, within and across languages.

\section{Methods}

\subsection{Models}

We consider seven diverse LLMs available on HuggingFace \cite{wolf2019huggingface}, ranging from 1.5 to 7.2 billion parameters (see Table \ref{tab:models}).

\begin{table}
\setlength{\tabcolsep}{3pt}
\centering
\resizebox{\linewidth}{!}{
\begin{tabular}{l l c c c r}
\toprule
Model & Authors \& Year & Param. & L & H & Lang. \\
\midrule
GPT2-XL & \citet{radford-2019} & 1.5B & 48 & 1600 & 1 \\
Llama-3.2-3B & \citet{meta-2024} & 3.2B & 28 & 3072 & 8 \\
Falcon3-3B-Base & \citetalias{tii-2024} & 3.2B & 22 & 3072 & 4 \\
Phi-4-mini-instruct & \citet{microsoft-2025} & 3.8B & 32 & 3072 & 23 \\
Gemma-3-4B-PT & \citet{deepmind-2025} & 4.3B & 34 & 2560 & 140+ \\
DeepSeek-LLM-7B-Base & \citet{deepseek-2024} & 6.9B & 30 & 4096 & 2 \\
Mistral-7B-v0.3 & \citet{mistral-2023} & 7.2B & 32 & 4096 & 1 \\
\bottomrule
\end{tabular}}
\caption{Models used in our experiments, sorted according to parameter count (Param.). For each model, we additionally report the number of layers (L), hidden size (H), and the number of languages included during training (Lang.)}
\label{tab:models}
\end{table}

\subsection{Linguistic materials}

Our primary focus is on the BLiMP benchmark \cite{warstadt-2020}, which contains sets of 1,000 minimal grammatical-ungrammatical sentence pairs for 67 syntactic phenomena, grouped into 12 higher-level categories based on a particular linguistic framework \cite{chomsky-1981}. The purpose of BLiMP is to assess LLMs' syntactic competence, as measured by their preference (higher probability, or, equivalently, lower surprisal) for grammatical sentences compared to their ungrammatical counterparts. The benchmark was constructed automatically and subsequently validated by human English speakers.

We additionally consider the English minimal pair materials from \citet{linzen-2016}, \citet{gulordava-2018}, and \citet{gauthier-2020}, as well as the non-English benchmarks: SLING (Chinese; \citealp{song-2022}), RuBLiMP (Russian; \citealp{taktasheva-2024}), and MultiBLiMP (multilingual; \citealp{jumelet-2025}).

\subsection{Unit localization approach}

Inspired by functional localization in the human brain (e.g., \citealt{fedorenko2010new}), we identify LLM units with the largest difference in activation magnitude between grammatical and corresponding ungrammatical sentences for each phenomenon. Following the approach in \citet{alkhamissi-2025}, we obtain (via last-token pooling) the LLM unit activations associated with each of the two conditions, take the absolute value of the activations, compute the Welch $t$-test between the two activation magnitude sets, and find the top-$k$ units with the greatest value of the $t$-statistic. In the main experiments, we search for top-1\% units (rounded down) over the residual stream (the outputs of all Transformer layers except for the embedding layer). We additionally show robustness of the key results to localizing 0.5\% (Appendix \ref{sec:appendix-0.5}) or 5\% (Appendix \ref{sec:appendix-5}) of units and conduct a more fine-grained search over the outputs of the attention and MLP modules (Appendix \ref{sec:appendix-finegrained}).

\section{Results}

\subsection{Do LLMs contain units consistently associated with—and causally important for—specific syntactic phenomena in English?}
\label{sec:english_identification_causal}

\begin{figure}[ht]
\includegraphics[width=\columnwidth]{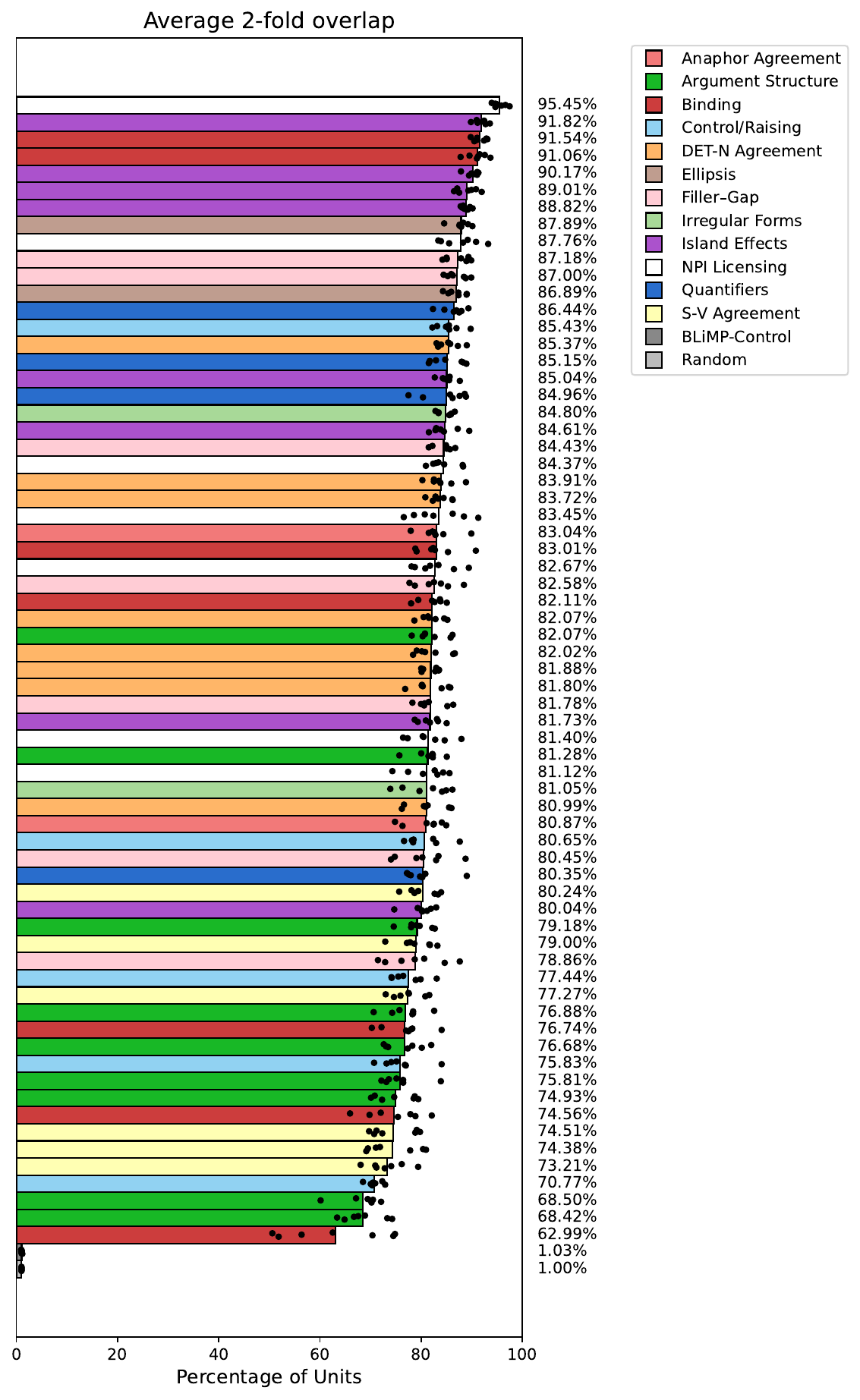}
\caption{\textbf{Consistency of LLM units engaged across sentence instances for each syntactic phenomenon.}
Bars show overlap between unit sets identified in two independent halves of the data (2-fold cross-validation) for each of the 67 BLiMP phenomena, averaged across seven models. Bars are sorted from highest to lowest overlap (with percent overlap shown on the right);  bar colors denote the category groupings in BLiMP and dots show individual models. Gray bars show two control conditions: \emph{Random} (analytical expected overlap) and \emph{BLiMP-Control} (randomized condition labels applied to grammatical sentences).}
\label{fig:cross-validation}
\end{figure}

\textbf{Consistency of unit recruitment across instances of the same phenomenon.}
We first asked: For a given syntactic phenomenon, do different sentence instances consistently recruit the same set of model units? To answer this question, we identified the top 1\% of model units most responsive to a given phenomenon and tested whether these units generalize across new minimal pairs that target the same phenomenon. For each phenomenon in the BLiMP benchmark, we performed 2-fold cross-validation: we split the 1,000 sentence pairs in half, performed localization independently on each set of 500 pairs, and recorded the overlap of the resulting units.
Figure \ref{fig:cross-validation} shows the average overlap in localized units across splits for each of the 67 phenomena. The overlaps are consistently high---ranging from 62.99\% to 95.45\%---indicating that these units respond reliably to a given phenomenon across different sentence instances. (These results are robust to 5-fold cross-validation, albeit with somewhat lower overlap scores---see Appendix \ref{sec:appendix-5-fold}.)

To situate these findings, we evaluated our overlap scores against two controls. First, we quantified chance-level overlap by computing the expected 2-fold intersection if 1\% of units were sampled uniformly at random in each fold---yielding an expected overlap of 0.01\% of all LLM units (\emph{Random} in Figure \ref{fig:cross-validation}). Second, we extracted only the grammatical sentences from each phenomenon (1,000 sentences), re-assigned condition labels (so 500 of them became ``ungrammatical'') to form 500 pseudo-minimal pairs, and separated the resulting pairs into two folds of 250 pairs each. We then averaged the cross-validated unit overlap across the 67 pseudo-phenomena. If the isolated units responded to non-syntactic regularities (e.g., BLiMP template effects) or were outliers with unusually high activation magnitude or variance \citep{he2024understanding}, pseudo-overlap would be high.
This control (\emph{BLiMP-Control}) also shows very low overlap (1.03\%), indicating that the consistency of identified syntactic units cannot be explained by trivial factors.

\begin{figure}[ht]
\centering
\includegraphics[width=\linewidth]{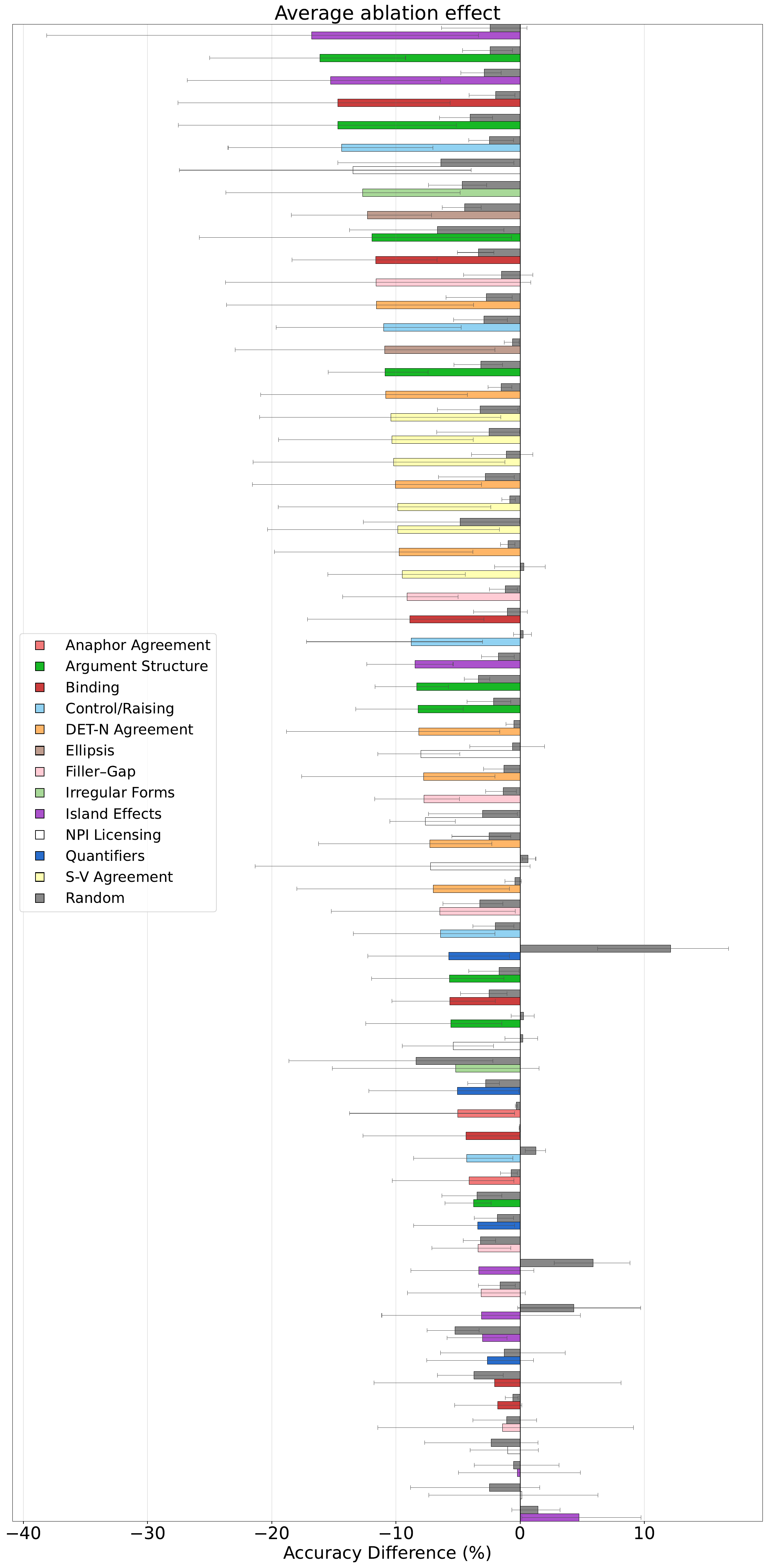}
\caption{\textbf{Performance impact of syntax-responsive unit ablation.} Each colored bar shows the average difference in accuracy between the top-unit-ablated (1\%) and unablated models for each of the 67 BLiMP phenomena, averaged across seven models. Bars are sorted from highest to lowest top-unit ablation performance drop. Gray bars show accuracy differences from ablating a random 1\% of units (averaged across seven models and four random seeds). Error bars denote 95\% confidence intervals over models.}
\label{fig:ablation}
\end{figure} 

Finally, to assess whether our findings extend beyond BLiMP—which is large but automatically generated---we investigated whether units localized on BLiMP generalize to other benchmarks targeting a subset of the syntactic phenomena (subject-verb agreement and filler-gap dependencies; \citealp{gauthier-2020,gulordava-2018,linzen-2016}). We observe significant overlap, even though the benchmarks vary in size, sentence length, and construction method (see Appendix \ref{sec:appendix-generalization}). 

\textbf{Causal involvement of phenomenon-relevant units in BLiMP performance.} Having identified LLM units that respond consistently to a given syntactic phenomenon, we asked: Do these syntax-responsive units causally support model performance on the grammaticality judgment task? We used half of the BLiMP sentence pairs for a given phenomenon (500 sentence pairs) for unit localization, zero-ablated the identified units, and evaluated the model's accuracy on the held-out half using the \texttt{minicons} package \cite{misra2022minicons} (accuracy is defined as the fraction of minimal pairs for which the model assigns higher probability to the grammatical sentence).
Figure \ref{fig:ablation} shows the resulting performance drops for each phenomenon: ablating the top-1\% of phenomenon-relevant units led to substantially larger performance drops than ablating a random 1\% of units (across four random seeds). 
Across phenomena and models, ablating phenomenon-relevant units reduced performance by an average of 7.61\%, compared with 1.68\% for random unit ablations. We replicated higher-than-random performance drops when ablating top-0.5\% and top-5\% of units (Appendices \ref{sec:appendix-0.5} and \ref{sec:appendix-5}) and when performing mean ablation instead of zero ablation (Appendix \ref{sec:appendix-mean-ablation}).


We observe substantial variability across models---our seven models show average ablation performance drops of 4.13\%, 3.92\%, 5.28\%, 5.54\%, 20.91\%, 0.76\%, and 12.72\%. The strongest ablation effects (20.91\% and 12.72\%) are observed in Gemma and Mistral respectively, which display clear and consistent accuracy drops following ablation across phenomena, suggesting that the localized units play a causal role in syntactic judgments. The lowest effect (0.76\%) is found in DeepSeek, which shows inconsistent effects (Appendix \ref{sec:appendix-ablation}).

We also observe variability across phenomena. To better understand this variability, we investigated whether phenomena for which the LLM units show higher cross-validation consistency are associated with larger ablation effects, but found no significant relationship (see Appendix \ref{sec:appendix-scatterplot}). 

\subsection{Do different syntactic phenomena recruit the same model units?}
\label{sec:english_cross_phenomenon}

\textbf{Unit overlaps within and across theoretically-defined syntactic categories.} 

Having shown that specific units within LLMs respond reliably to---and are causally important for---individual syntactic phenomena, we asked whether the identified units overlap across phenomena. In other words, do different syntactic phenomena recruit the same model units, or do they instead recruit distinct model components?

\begin{figure}[ht]
\includegraphics[width=\columnwidth]{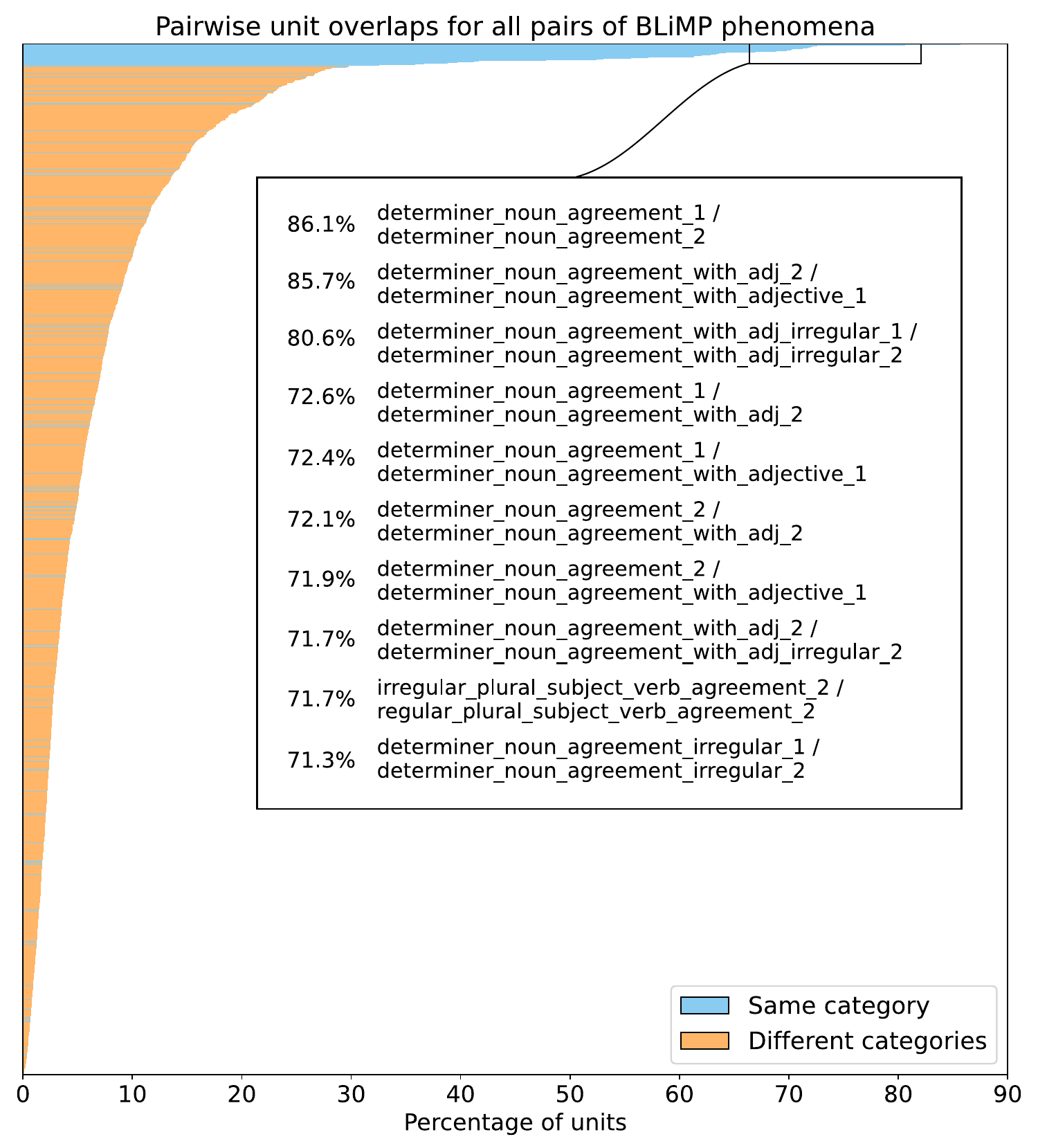}
\caption{\textbf{Unit overlaps for all 2,211 pairs of BLiMP phenomena, averaged across seven models.} Blue bars indicate unit overlaps for pairs of phenomena within the same syntactic category; orange bars indicate overlaps for pairs of phenomena across categories. The inset shows the pairs with the highest overlaps.}
\label{fig:cross-overlap-histogram}
\end{figure}

To address this question, we performed localization independently for every BLiMP phenomenon (using the full set of 1,000 sentences), and then measured overlaps for the resulting unit sets across all phenomenon pairs ($\frac{67 \times 66}{2} = 2,211$ pairs). Figure \ref{fig:cross-overlap-histogram} shows the overlaps for all pairs, ordered by overlap size. Some pairs show high overlap: up to 86.05\% ({\ttfamily determiner\_noun\_agreement\_1} and {\ttfamily determiner\_noun\_agreement\_2}, which both test determiner-noun number agreement, but in the former, the grammaticality contrast is achieved through changing the number of the noun, and in the latter, the determiner). Other pairs show no or very little overlap (e.g., {\ttfamily dist\-ractor\_agreement\_relational\_noun} and {\ttfamily left\-\_branch\_island\_echo\_question} have 0.05\% overlap). Interestingly, no unit in any model is in the top-1\% set for all 67 BLiMP phenomena, with the most syntax-general unit being shared among 52 phenomena in the Gemma model.

As noted above, the phenomena in BLiMP are grouped into 12 categories. So we next evaluated whether pairs of phenomena assigned to the same category show higher inter-phenomenon overlaps than pairs straddling category boundaries. Note that the grouping in BLiMP is theory-based, not evidence-based, so this examination is the first attempt to empirically evaluate the grouping of syntactic phenomena into higher-order categories.

\begin{figure}
\includegraphics[width=\columnwidth]{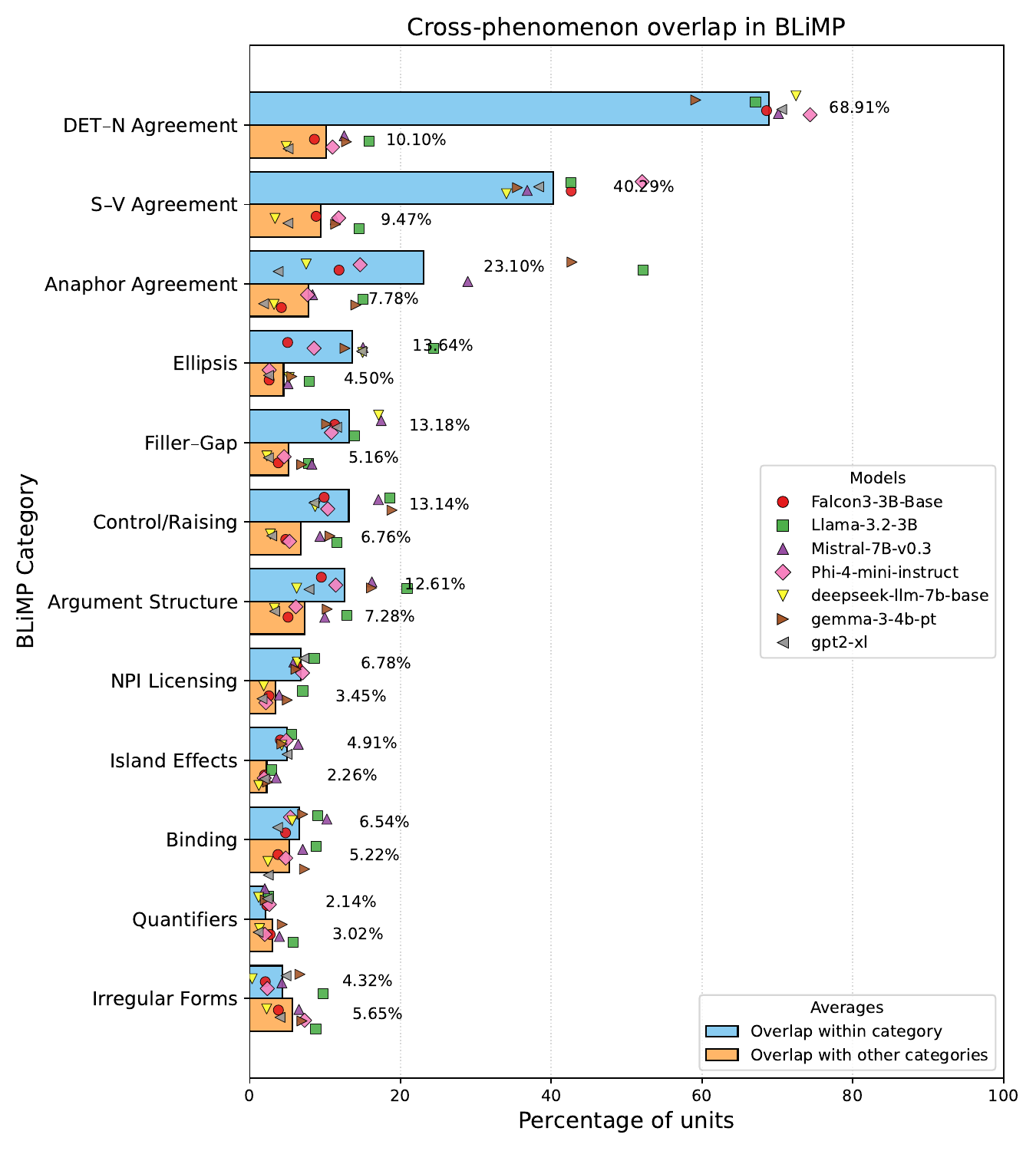}
\caption{\textbf{Within-category and cross-category overlaps in BLiMP.} Each blue bar shows the average (across all pairs of phenomena within a category) intersection of localized unit sets (as a percentage out of the 1\% target set; averaged across models). Each corresponding orange bar shows the average intersection across all pairs of phenomena belonging to distinct categories. The categories are ordered by the difference between the within-category and cross-category average overlap. Markers denote individual models.}
\label{fig:cross-overlap}
\end{figure}

The results are shown in Figure \ref{fig:cross-overlap}: across most phenomena, the within-category overlaps are reliably higher than the cross-category overlaps ($t=2.07$, $p=0.0312$ via a two-sample Welch $t$-test). Interestingly, this pattern appears to be driven by the three agreement categories: determiner-noun agreement, subject-verb agreement, and anaphor agreement (average within-category overlaps: 68.91\%, 40.29\%, and 23.10\%, respectively). Other categories (e.g., ellipsis and argument structure) show much smaller differences between the within-category vs. cross-category overlaps. And some categories (e.g., quantifiers and irregular forms) show minimal within-category overlap (in the presence of high within-phenomenon overlap; Figure \ref{fig:cross-validation}), which suggests that these phenomena do not form meaningful categories within LLMs and instead rely on phenomenon-specific units. See Appendixes \ref{sec:appendix-0.5} and \ref{sec:appendix-5} for results at 0.5\% and 5\% localization thresholds, which show the same patterns.

\begin{figure}
\includegraphics[width=\columnwidth]{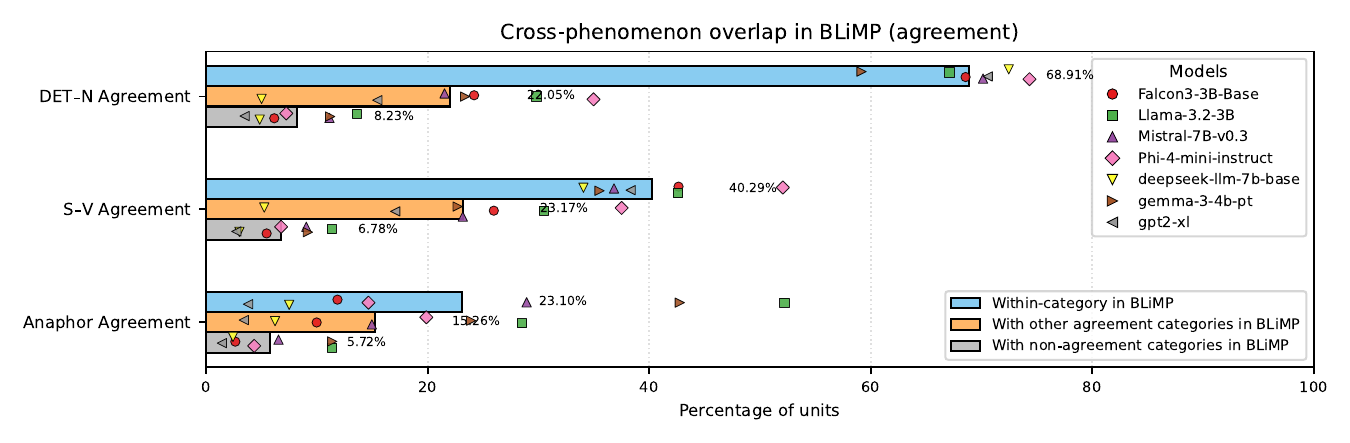}
\caption{\textbf{Overlap of localized units within and across agreement categories.} Bars show average overlap (as a percentage of top-1\% units; averaged across models) between pairs of phenomena: (i) within each agreement category (blue; same data as in Figure \ref{fig:cross-overlap}), (ii) across agreement categories (orange), and (iii) between agreement and non-agreement phenomena (gray).}
\label{fig:cross-overlap-agreement}
\end{figure}

\textbf{Unit overlap among different types of agreement phenomena.} Because the three most consistent categories were all related to syntactic agreement (determiner-noun, subject-verb, and anaphor agreement), we next investigated those categories in greater detail by computing overlaps among them, compared to overlaps with phenomena from non-agreement categories. Figure \ref{fig:cross-overlap-agreement} shows that for each agreement category, overlap with phenomena from other agreement categories is substantially higher (ranging from 15.26\% to 23.17\%) than with non-agreement phenomena (5.72\% to 8.23\%). This result suggests that different kinds of agreement draw on shared resources within LLMs.

To test whether the units we have identified as being consistently engaged by syntactic agreement may also be sensitive to other kinds of linguistic violations, we developed an additional control condition (beyond \emph{Random} and \emph{BLiMP-Control} in Section~\ref{sec:english_identification_causal}), which we dubbed \emph{BLiMP-Lex}, by replacing either nouns or verbs in the grammatical BLiMP sentences with length- and frequency-matched alternatives. We manually reviewed the substitutions to ensure that they only affected sentence meaning without introducing syntactic (e.g., subcategorization) violations. We then localized units using these noun- and verb-substituted materials and measured their overlap with the syntactic agreement units. We obtained a syntactic-lexical overlap score of 13.33\%, which is somewhat lower than the syntactic overlaps across categories (Figure \ref{fig:cross-overlap-agreement})---indicating that the agreement units primarily capture syntactic, not semantic, distinctions (see details in Appendix \ref{appendix-blimp-lex}).

\subsection{Examination of cross-phenomenon relationships in languages other than English}  \label{sec:multilingual}

Building on the English findings, we next asked whether our main results concerning the cross-phenomenon relationships generalize to other languages. To do so, we used two additional benchmarks: RuBLiMP for Russian (an Indo-European language; \citealp{taktasheva-2024}) and SLING for Chinese (a Sino-Tibetan language; \citealp{song-2022}). Both benchmarks are conceptually and structurally similar to BLiMP: RuBLiMP contains 45 phenomena with 1,000 minimal pairs each, grouped into 12 theoretically-defined syntactic categories, while SLING contains 38 phenomena with 1,000 pairs grouped into 9 categories. We conducted the analyses with the Gemma model because it supports both languages.

\begin{figure*}
  \centering
  
  \begin{subfigure}
    \centering
    \includegraphics[width=.49\linewidth]{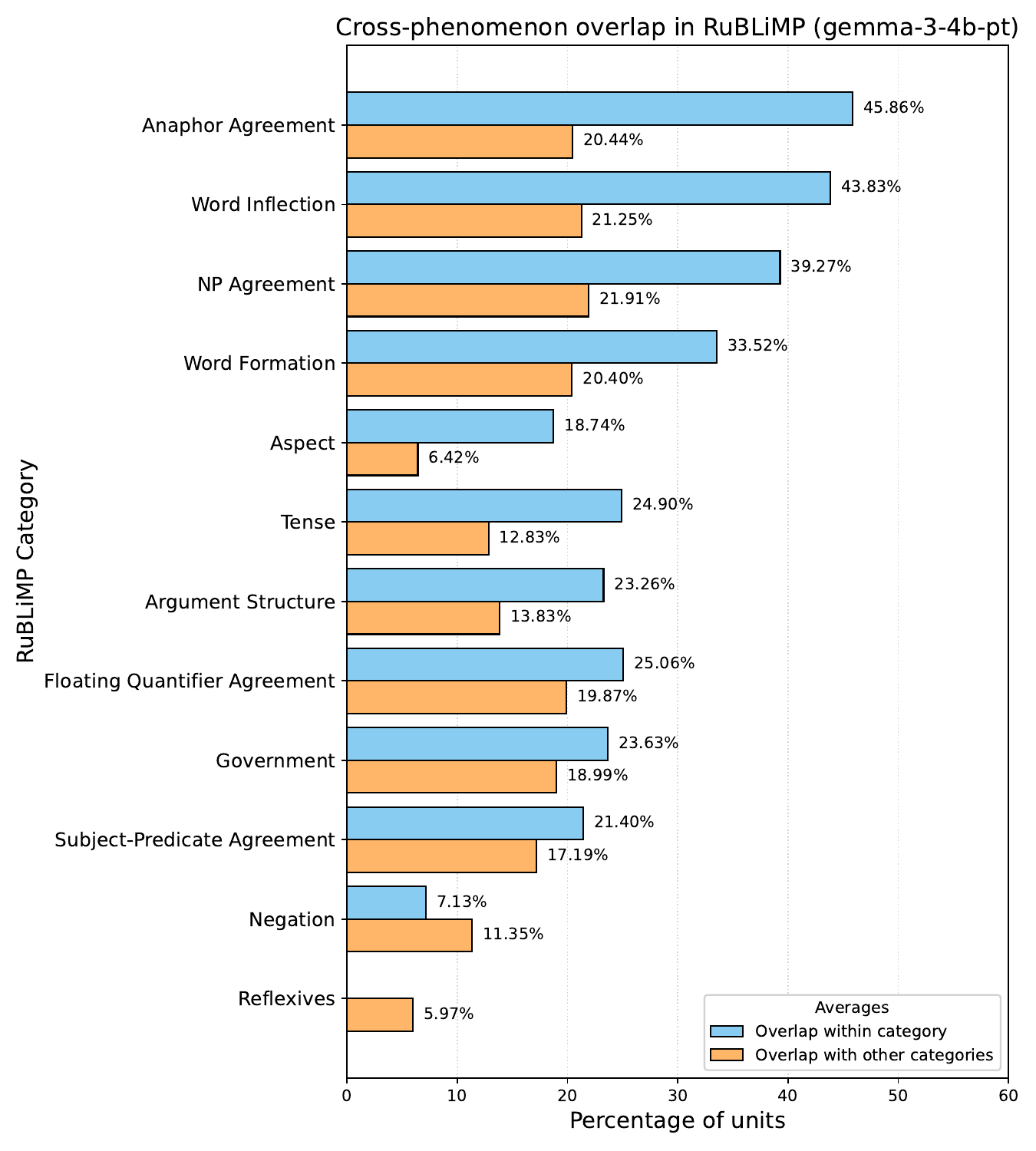}
  \end{subfigure}
  \begin{subfigure}
    \centering
    \includegraphics[width=.49\linewidth]{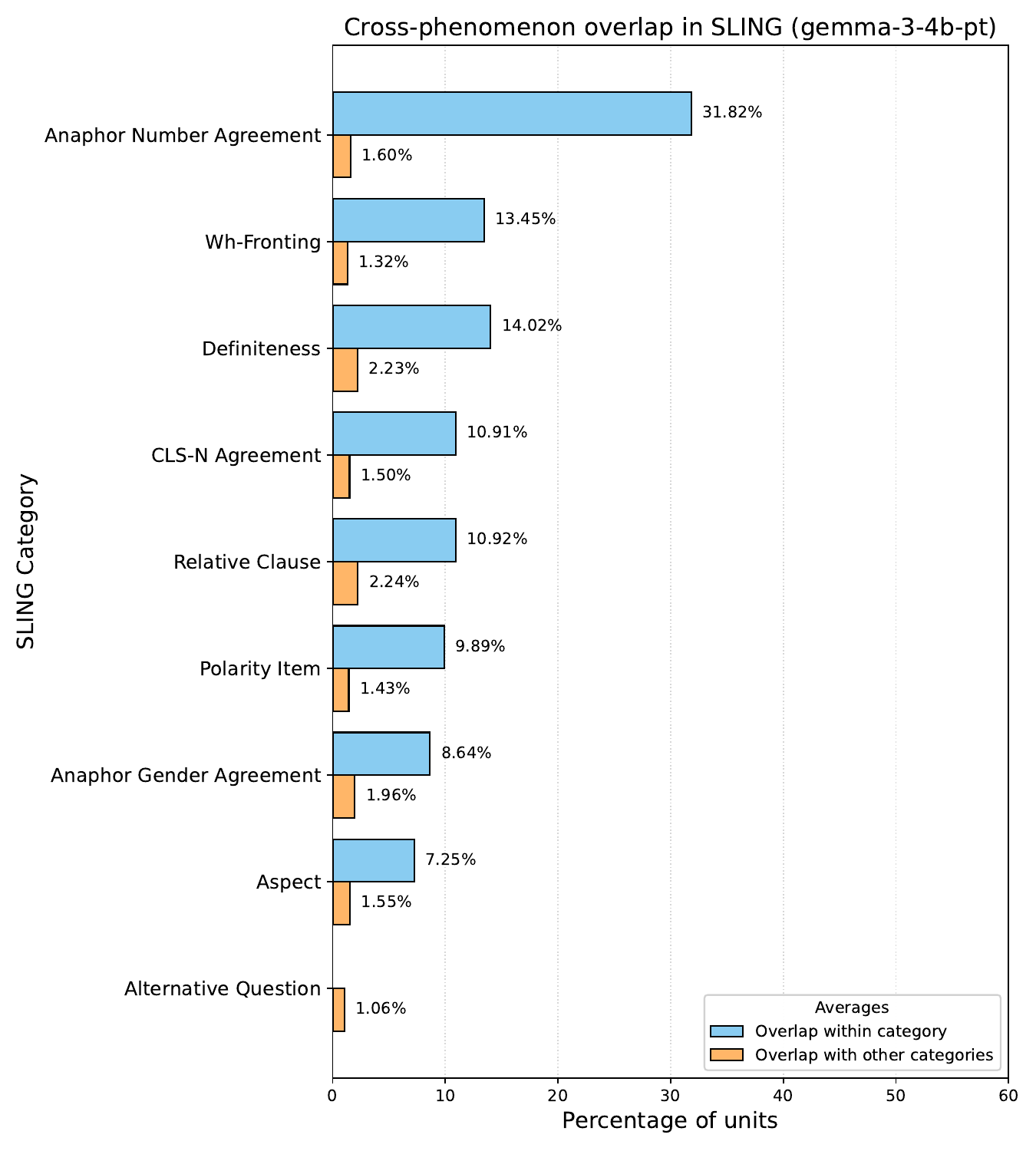}
  \end{subfigure}

  \caption{\textbf{Within-category and cross-category overlaps in Russian (RuBLiMP) and Chinese (SLING).} The bars show average (across phenomena) within-category and cross-category overlaps for Russian (left) and Chinese (right), both using the Gemma model. The categories are ordered by the difference between the within-category and cross-category average overlap.}
  \label{fig:cross-overlap-rublimp-sling}
\end{figure*}

\textbf{Unit overlaps within and across syntactic categories in Russian and Chinese.} We first replicated the finding that individual phenomena recruit consistent sets of units: both Russian and Chinese exhibit high 2-fold consistency (mean 2-fold overlap for Gemma is 65.97\% for Russian and 81.60\% for Chinese, compared to 82.52\% for English; see Appendix \ref{sec:appendix-cv-gemma}). Critically, similar to English, overlap for pairs of phenomena within a category tends to be higher than across categories. Moreover, as in English, the categories showing the largest differences between the within- and cross-category comparisons tend to be the agreement categories (Russian: anaphor and noun phrase agreement are in the top three; Chinese: anaphor number agreement is the top category), although some agreement categories do not show pronounced differences (Figure \ref{fig:cross-overlap-rublimp-sling}).

\textbf{Unit overlaps for agreement phenomena across languages.} In addition to allowing for investigation of linguistic phenomena in different languages, multilingual models allow for testing for potential overlap in linguistic computations \textit{across} languages. We next examined whether the previous findings of shared syntactic representations across languages \citep{chi-etal-2020-finding, stanczak-etal-2022-neurons} hold for agreement-related units (see also \citealp{mueller-etal-2022-causal, devarda2024syntax}).

\textbf{Unit overlap for agreement versus other phenomena between English and Russian/Chinese.} Given that agreement has emerged as a domain with high within-category consistency (Section~\ref{sec:english_cross_phenomenon}), we focus on the agreement phenomena, testing whether the agreement-relevant units identified in Russian and Chinese overlap with the agreement-relevant units in English, compared to units relevant to other phenomena.

In Russian, verbs have suffixes that encode the number and person (for present and future tense forms) or gender (for past tense forms) of the subject. Adjectives and determiners have suffixes that agree in number, gender, and case with the noun they modify. RuBLiMP contains phenomena focused on number, person, and gender agreement between subjects and predicates; number and gender agreement between anaphors and their antecedents; number, gender, and case agreement between adjectives and nouns.

In Chinese, numerals and determiners are linked to nouns via classifiers, which must agree with the noun semantically (the choice of the classifier depends on factors like animacy and shape). Chinese also has a singular/plural distinction in pronouns and a gender distinction between ``he'' and ``she'' (only in writing). SLING contains phenomena focused on classifier-noun agreement as well as the number and gender agreement between anaphors and their antecedents.

\begin{figure}
  \centering
  
  \begin{subfigure}
    \centering
    \includegraphics[width=\columnwidth]{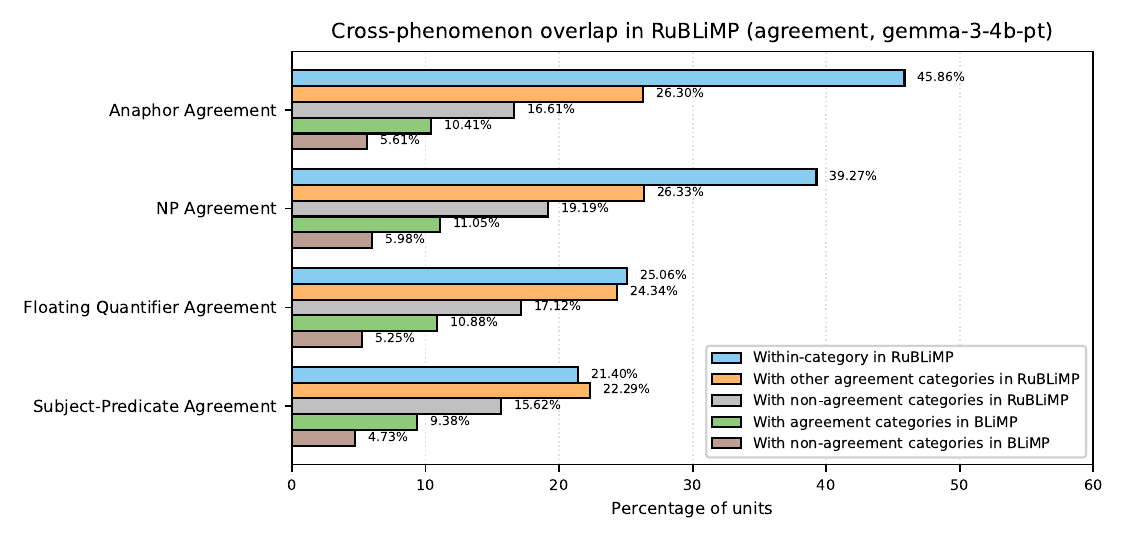}
  \end{subfigure}
  
  \begin{subfigure}
    \centering
    \includegraphics[width=\columnwidth]{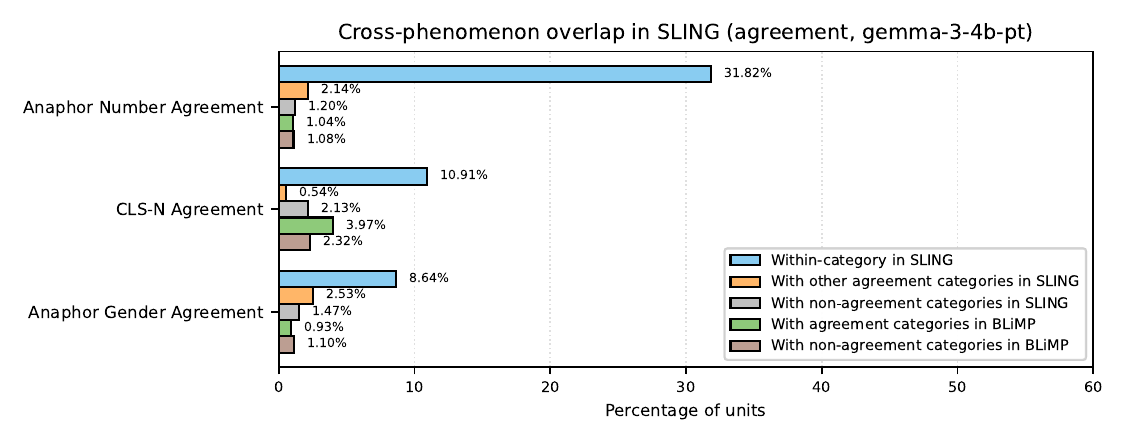}
  \end{subfigure}

  \caption{\textbf{Cross-lingual overlap of localized units for the agreement categories.} For RuBLiMP (top) and SLING (bottom), bars show the average overlaps (as a percentage of the 1\% top units; both use the Gemma model): (i) within each agreement category (blue); (ii) across different agreement categories within the same language (orange); (iii) between agreement and non-agreement phenomena within the same language (gray); (iv) with the agreement categories in English (BLiMP) (green); (v) with the non-agreement categories in English (BLiMP) (brown).}
  \label{fig:cross-overlap-agreement-rublimp-sling}
\end{figure}

Indeed, agreement-relevant units generalize across languages, especially between Russian and English (average overlap across pairs of agreement phenomena = 9.99\%; green bars in Figure \ref{fig:cross-overlap-agreement-rublimp-sling}), but also between Chinese and English, albeit to a lesser extent (average overlap = 2.22\%); the overlap with non-agreement categories in English is shown in brown bars. To contextualize this overlap, we compare it with the overlaps obtained within a language (i) among the phenomena within the relevant agreement category as the upper bound (e.g., anaphor agreement; blue bars), (ii) with other agreement categories (e.g., anaphor agreement and noun phrase agreement; orange bars), and (iii) with non-agreement categories (gray bars). The cross-language overlap is substantially lower than the overlap within a language, indicating that although some agreement-relevant units are shared across languages, many remain language-specific.

\textbf{Unit overlap for subject-verb agreement as a function of language similarity.} In the previous section, the overlap between the agreement-relevant units was higher for English and Russian, compared to English and Chinese. One explanation is that English and Russian are more syntactically similar. To test the hypothesis that syntactic similarity influences unit overlap, we used the MultiBLiMP benchmark \cite{jumelet-2025}, which contains subject-verb number agreement sentence pairs (denoted SV-\#) for 90 languages. Using the Gemma model, we first investigated the range of overlaps across all $\binom{90}{2} = 4,005$ language pairs, finding variability---from 71.84\% for the most overlapping language pair (Serbo-Croatian and Czech) to no overlap (e.g., Tamil and Breton).

\begin{figure}
\includegraphics[width=\linewidth]{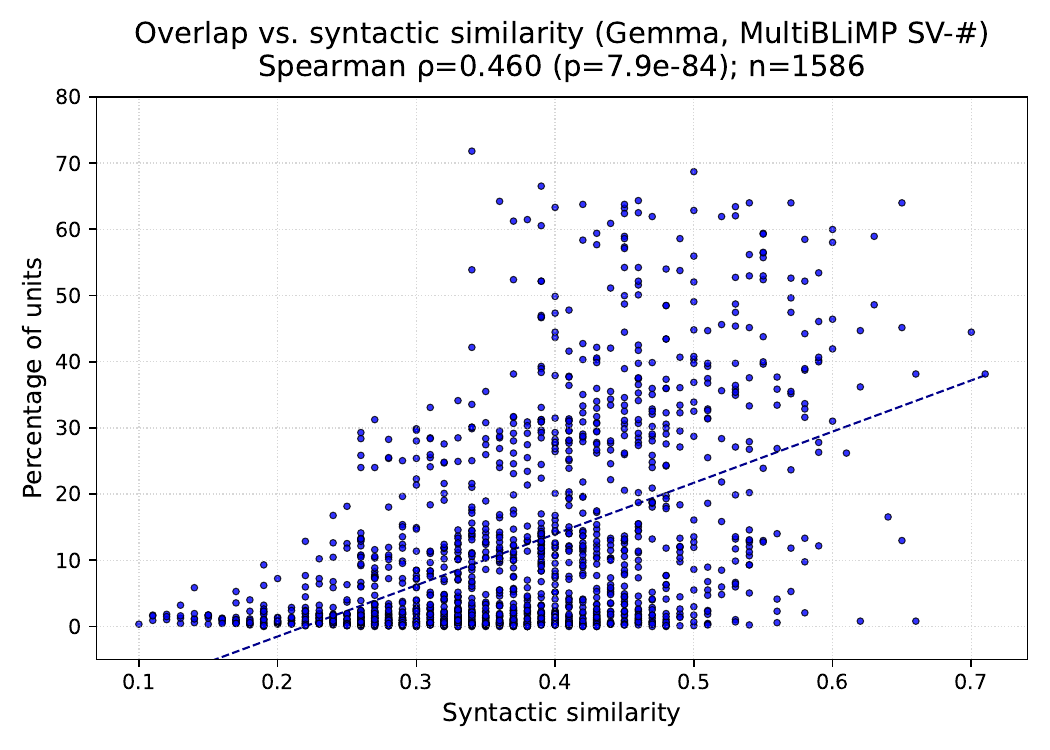}
\caption{\textbf{Relationship between language relatedness (syntactic similarity) and unit overlap, using the SV-\# split from MultiBLiMP across 57 languages (1,586 pairs).} Each dot represents a language pair; the $x$-coordinate is the syntactic similarity between the two languages in the pair, and the $y$-coordinate is the unit overlap. The dotted line shows the line of best fit.}
\label{fig:multiblimp}
\end{figure}

Next, we examined how unit overlap varies with syntactic similarity across languages. Syntactic similarity was computed using \texttt{lang2vec} \citep{littell2017uriel}, which provides vectors encoding each language's syntactic features \cite{haspelmath-2005, collins-2011, campbell-2008} and precomputed cosine distances, which we transform to similarity as one minus distance. We focused on the languages that (i) are included in MultiBLiMP SV-\# and (ii) have complete feature vectors without missing features (57 languages and 1,596 pairwise similarity values). We excluded 10 language pairs with spurious maximal similarities despite non-identical vectors (likely placeholders from an incomplete update), resulting in a total of 1,586 language pairs. Figure \ref{fig:multiblimp} reveals that syntactically similar languages indeed have higher unit overlaps. This finding suggests that syntactic representations are organized in a way that reflects cross-linguistic similarity: the closer two languages are syntactically, the more they share subject-verb number agreement units.

\section{Discussion}

We have shown that LLMs contain units that are consistently recruited for individual syntactic phenomena, and that different types of syntactic agreement draw on overlapping subsets of those units, indicating that agreement constitutes a meaningful category within LLMs (as independently demonstrated in English, Chinese, and Russian). These agreement-selective units also show non-trivial overlap across languages, pointing to a shared pool of units that support agreement-related computations. Moreover, the number of shared units increases with syntactic similarity: agreement-related units overlap more between English and Russian than between English and Chinese (a finding we extended to 57 languages).

These results have several implications. First, they argue against the idea that LLMs contain generic syntax-processing units, i.e., units that support syntactic computations across diverse phenomena \citep[cf.][]{boguraev2025causal}. Instead, distinct model units represent different aspects of grammatical knowledge, even for phenomena that have been previously grouped on theoretical grounds \citep{chomsky-1981}.
However, a striking exception to these phenomenon-idiosyncratic units is agreement-related phenomena. In English, Russian, and Chinese, the most coherent categories---those with high within-category and low cross-category overlaps---are agreement categories. Moreover, different agreement categories overlap with each other, suggesting a shared functional substrate for syntactic agreement in LLMs.
Another implication has to do with cross-linguistic unit overlap in multilingual models. The fact that more syntactically similar languages exhibit greater overlap suggests that multilingual models organize syntactic representations in a way that reflects cross-linguistic similarity. 

Finally, these findings have implications for cognitive science and neuroscience: the patterns of cross-phenomenon overlap in LLMs motivate investigations of whether humans, for example, show greater neural overlap for some phenomena than others, and whether bilingual or multilingual speakers exhibit more similar syntactic processing for more closely related languages. Although many differences exist between human brains and LLMs, LLMs offer a powerful system for deriving hypotheses about the neural organization of linguistic knowledge \citep[e.g.][]{jain2024computational,tuckute2024language}.

\section*{Limitations}

Limitations of our work include the reliance on top-$n$\% unit localization, which may overlook more distributed syntactic representations. Our analyses are conducted in the models' native activation basis; alternate representations might reveal structure that is not captured here (e.g., in non-basis-aligned subspaces; \citealp{wu2023interpretability, geiger2024finding}; see \citealp{mueller2024quest} for an overview). Additionally, our analyses are primarily correlational: although we do demonstrate the causal involvement of the target units in Section~\ref{sec:english_identification_causal} and Appendix \ref{sec:appendix-ablation}, further causal interventions and circuit-level analyses will be needed to evaluate the importance of shared agreement units to model performance on agreement phenomena. Another limitation is that we have focused on benchmarks that depend on comparing grammatical sentences to their ungrammatical counterparts. However, grammatical knowledge and processing go beyond detecting syntactic violations.

In the future, we plan to apply a similar approach to other syntactic phenomena---including sentences that use infrequent structures, contain ambiguities, or long-distance dependencies---to develop a more comprehensive understanding of how grammatical knowledge and processing are instantiated in LLMs. Moreover, future work could extend these experiments to other languages for which BLiMP-like minimal pair benchmarks have recently been made available, such as Dutch (BLiMP-NL; \citealp{sujikerbujik-2025}) or Turkish (TurBLiMP; \citealp{başar-2025}). Another rich avenue for future work is understanding the relationship between syntax and semantics in model representations. We took an initial step toward this goal with \emph{BLiMP-Lex}, demonstrating that the localized agreement units respond more reliably to syntactic than semantic violations, but disentangling the two remains challenging \citep{huang2021disentangling,reddy-2021,zhang-2022}. 

\section*{Acknowledgements}
We acknowledge support from MIT's McGovern Institute for Brain Research and Siegel Family Quest for Intelligence.
This work has been made possible in part by a gift from the Chan Zuckerberg Initiative Foundation to establish the Kempner Institute at Harvard University. A.G.d.V. was supported by the K. Lisa Yang ICoN Center Fellowship.

\FloatBarrier

\bibliography{custom}

\appendix

\section{Analyses localizing top-0.5\% of model units}
\label{sec:appendix-0.5}

\begin{figure}[H]
\includegraphics[width=0.85\linewidth]{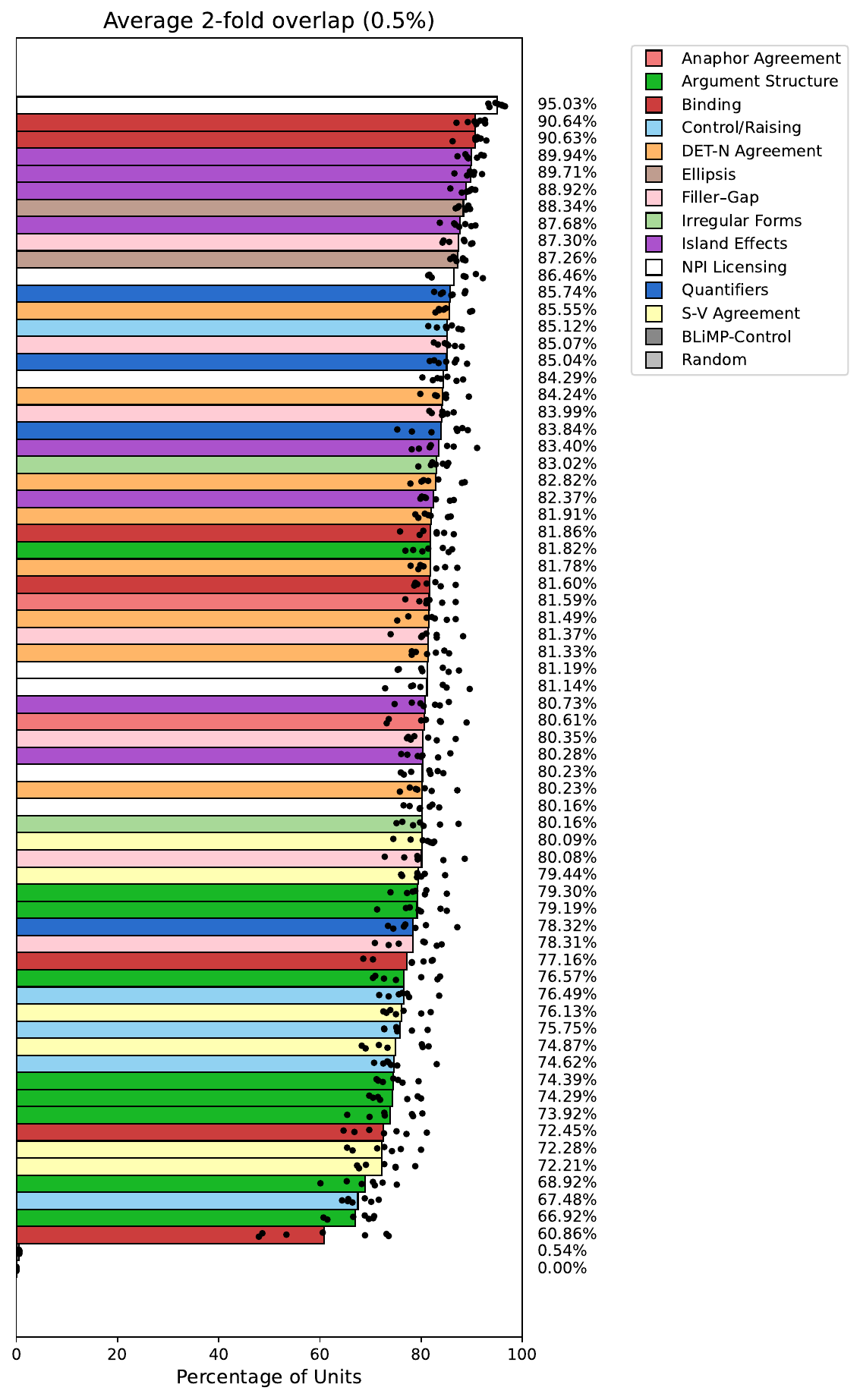}
\caption{Cross-validation analysis (2-fold) targeting 0.5\% of units.}
\label{fig:cv-0.5}
\end{figure}

\begin{figure}[H]
\includegraphics[width=0.72\columnwidth]{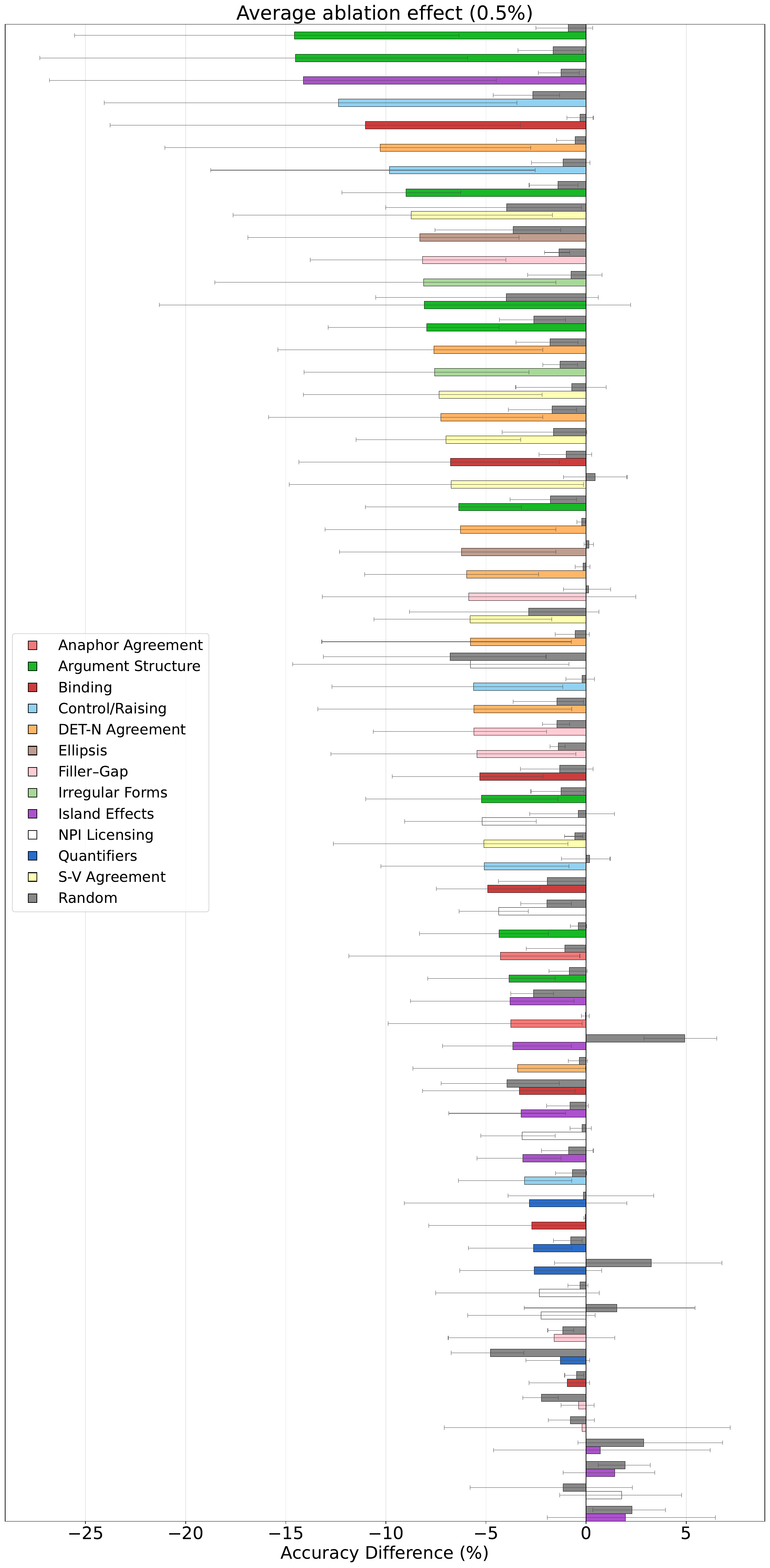}
\caption{Ablation analysis targeting 0.5\% of units.}
\label{fig:ablation-0.5}
\end{figure}

\begin{figure}[H]
\includegraphics[width=\columnwidth]{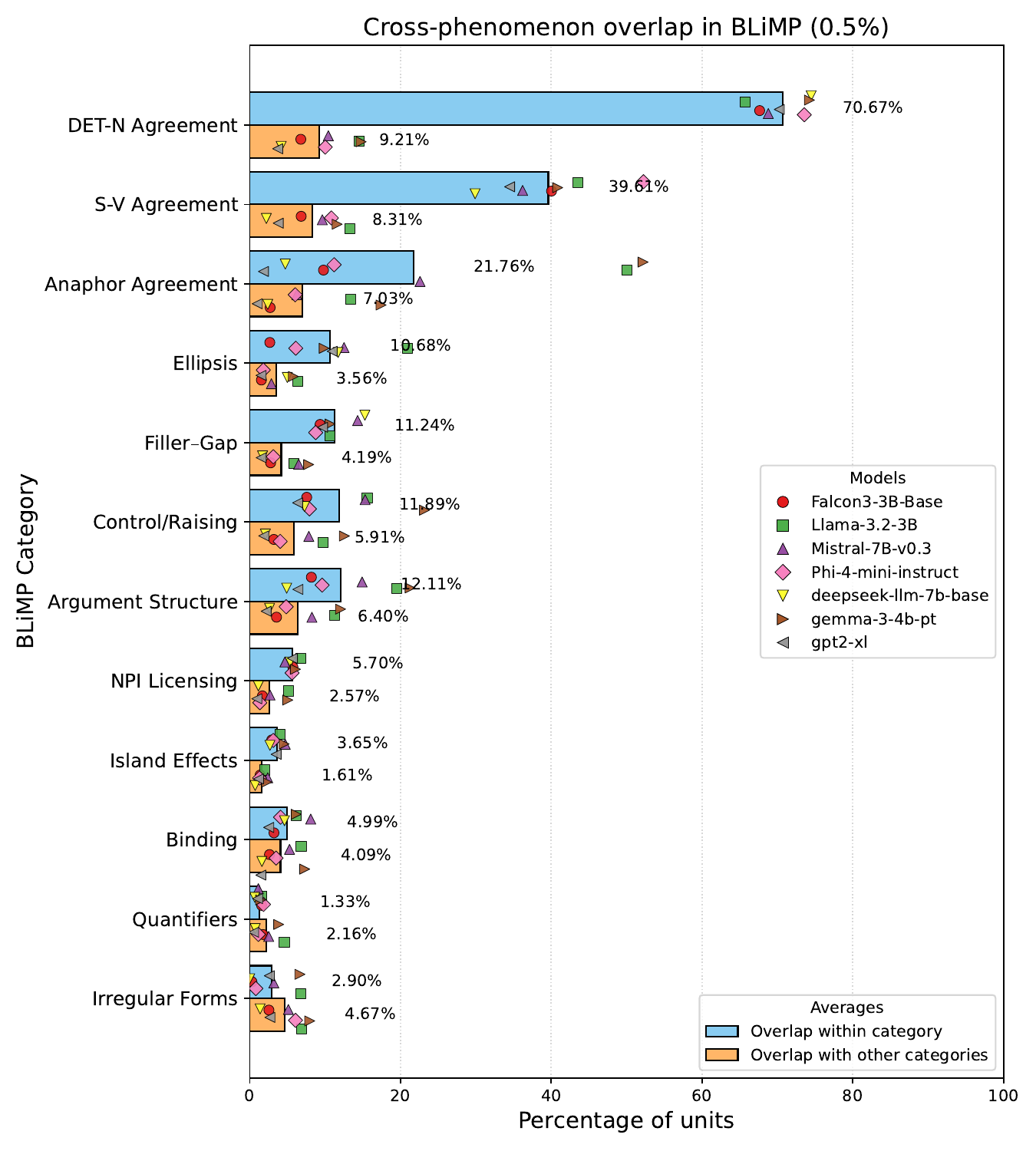}
\caption{Cross-phenomenon overlap analysis targeting 0.5\% of units.}
\label{fig:cross-overlap-0.5}
\end{figure}

We repeated the cross-validation and ablation experiments from Section~\ref{sec:english_identification_causal} and the cross-phenomenon overlap experiment from Section~\ref{sec:english_cross_phenomenon}, localizing the top-0.5\% of model units instead of top-1\%. Figure \ref{fig:cv-0.5} shows the 2-fold cross-validation result, Figure \ref{fig:ablation-0.5} shows the ablation result, and Figure \ref{fig:cross-overlap-0.5} shows the cross-phenomenon overlap result---all three are similar to the corresponding results targeting 1\% of units. The average ablation effect is 5.34\% (compared to ablating a random 0.5\% of units across four seeds: 0.97\%), demonstrating causal involvement of the localized units in BLiMP performance.

\section{Analyses localizing top-5\% of model units}
\label{sec:appendix-5}

\begin{figure}[H]
\includegraphics[width=\columnwidth]{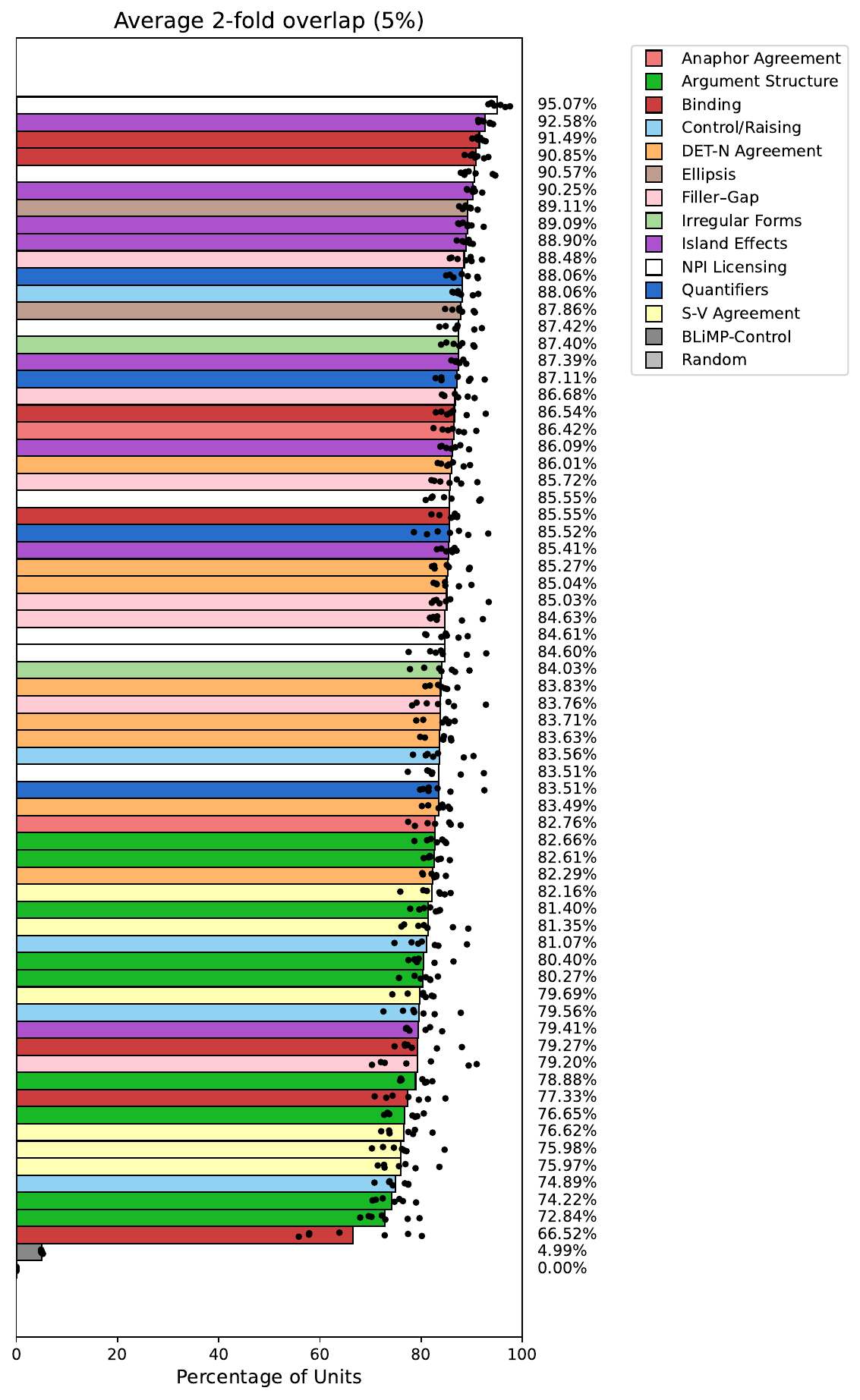}
\caption{Cross-validation analysis (2-fold) targeting 5\% of units.}
\label{fig:cv-5}
\end{figure}

\begin{figure}[H]
\includegraphics[width=0.86\columnwidth]{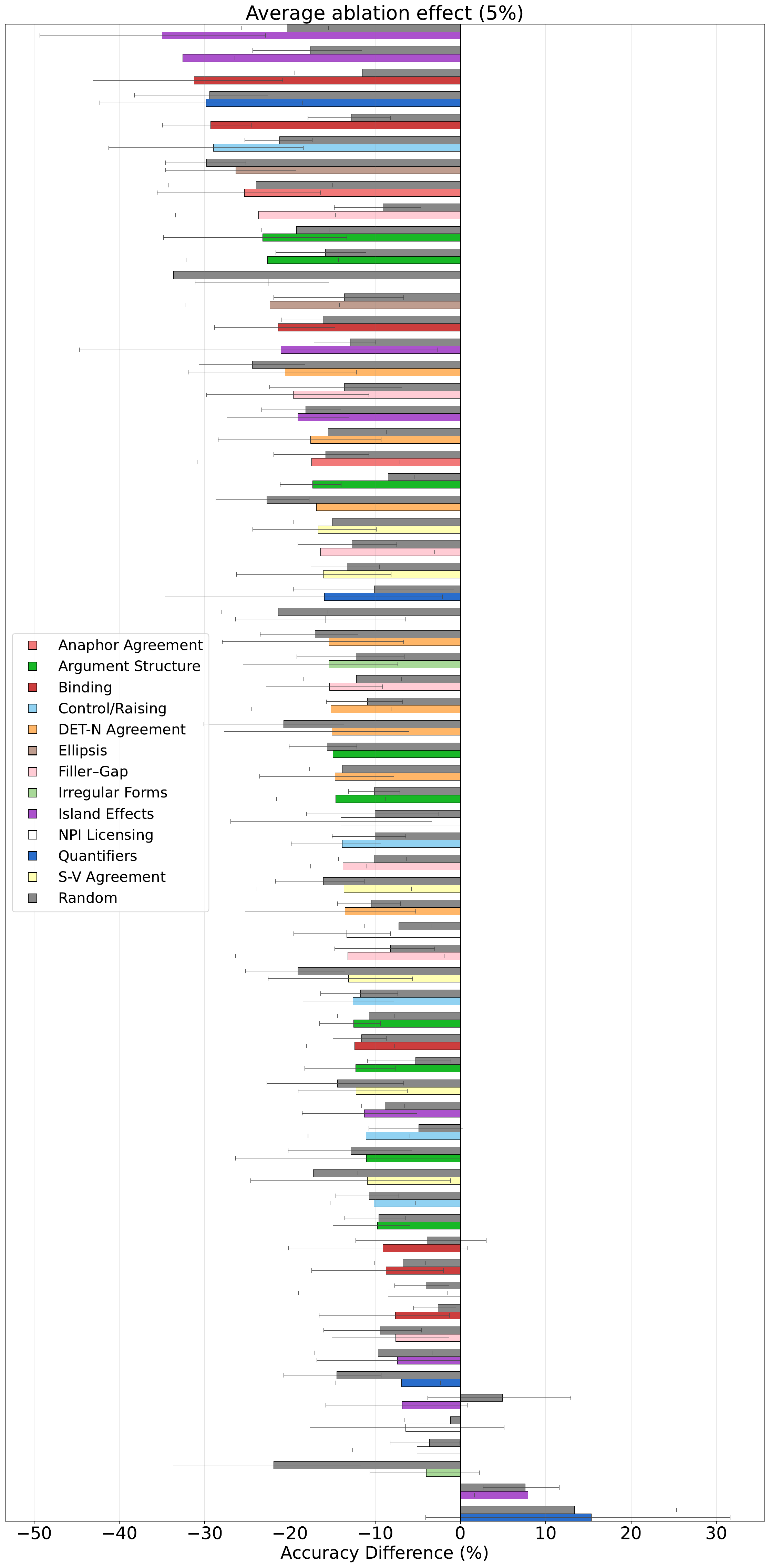}
\caption{Ablation analysis targeting 5\% of units.}
\label{fig:ablation-5}
\end{figure}

\begin{figure}[H]
\includegraphics[width=\columnwidth]{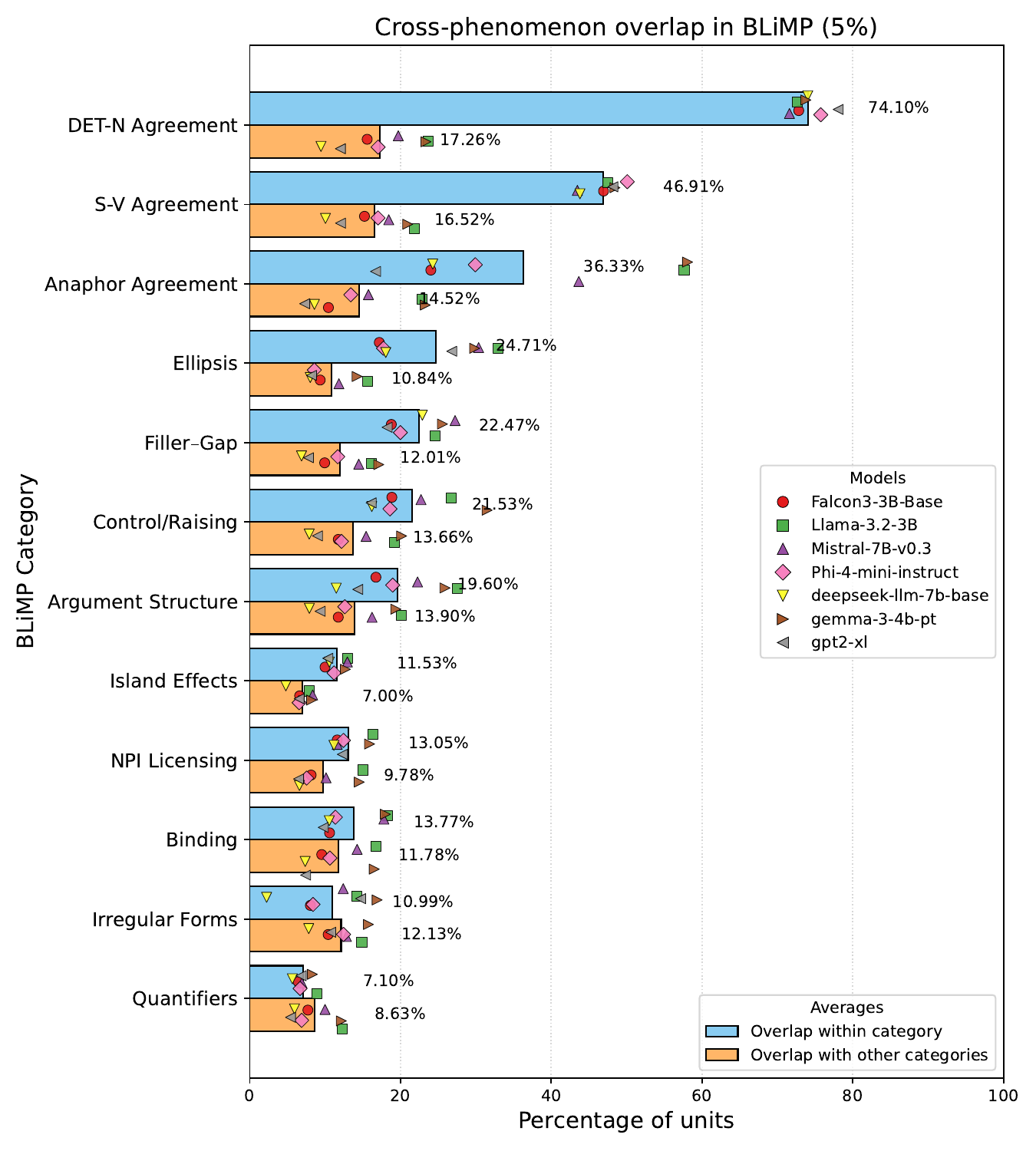}
\caption{Cross-phenomenon overlap analysis targeting 5\% of units. }
\label{fig:cross-overlap-5}
\end{figure}

We also repeated the same experiments targeting top-5\% of units. Figure \ref{fig:cv-5} shows the 2-fold cross-validation result, Figure \ref{fig:ablation-5} shows the ablation result, and Figure \ref{fig:cross-overlap-5} shows the cross-phenomenon overlap result. The average ablation effect is 15.17\% (compared to random-unit ablation across four seeds: 12.72\%).
For the cross-phenomenon overlap experiment (Figure \ref{fig:cross-overlap-5}), the qualitative findings are consistent with those in Figure \ref{fig:cross-overlap}, with particularly high within-category overlaps for the agreement phenomena.

\section{Analyses considering the outputs of attention and MLP modules}
\label{sec:appendix-finegrained}

\begin{figure}[H]
\includegraphics[width=\columnwidth]{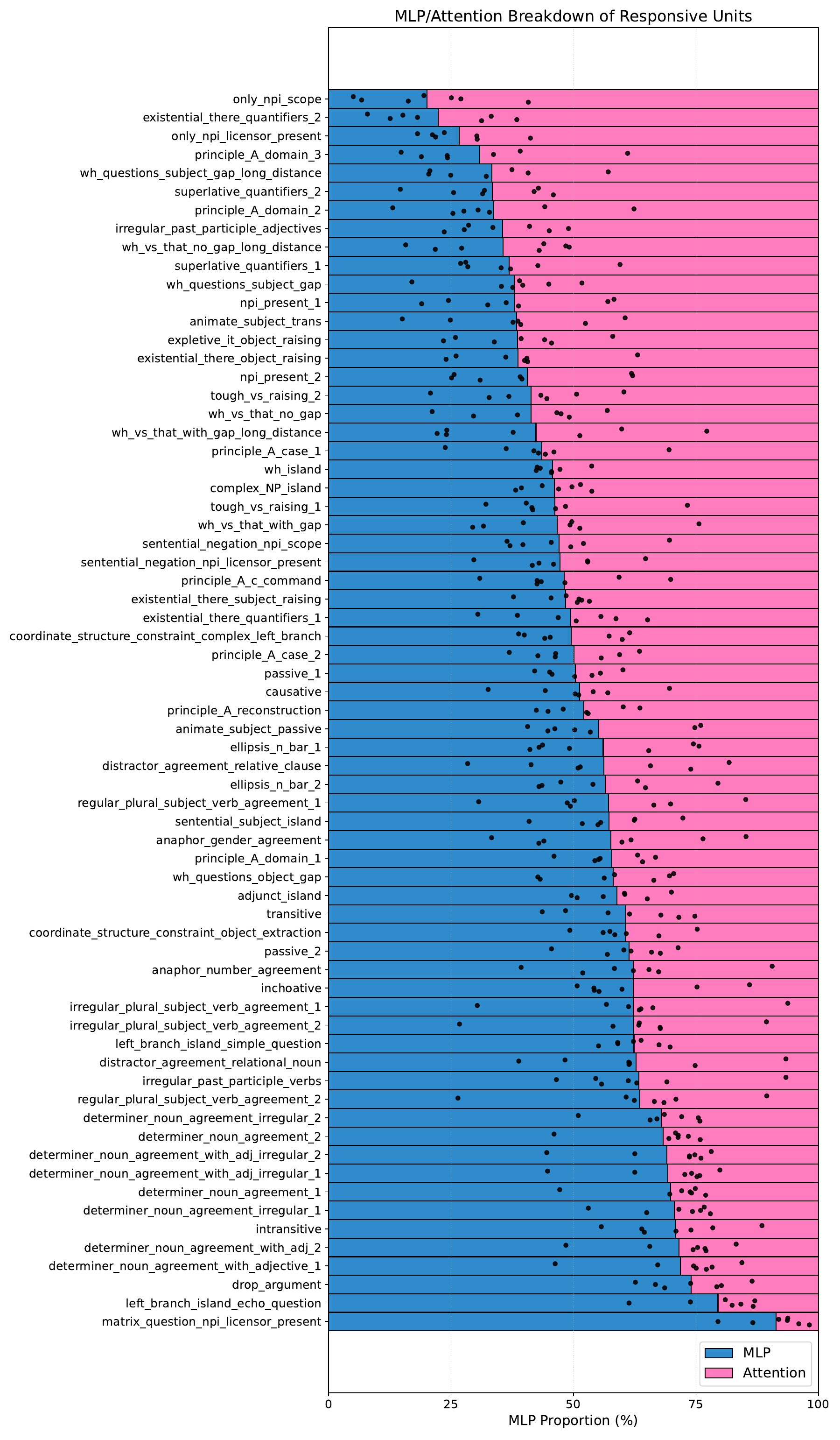}
\caption{\textbf{Breakdown of syntax-responsive units by module type:} MLP (blue) / attention (pink).}
\label{fig:mlp-attn}
\end{figure}

\begin{figure}[H]
\centering
\includegraphics[width=\columnwidth]{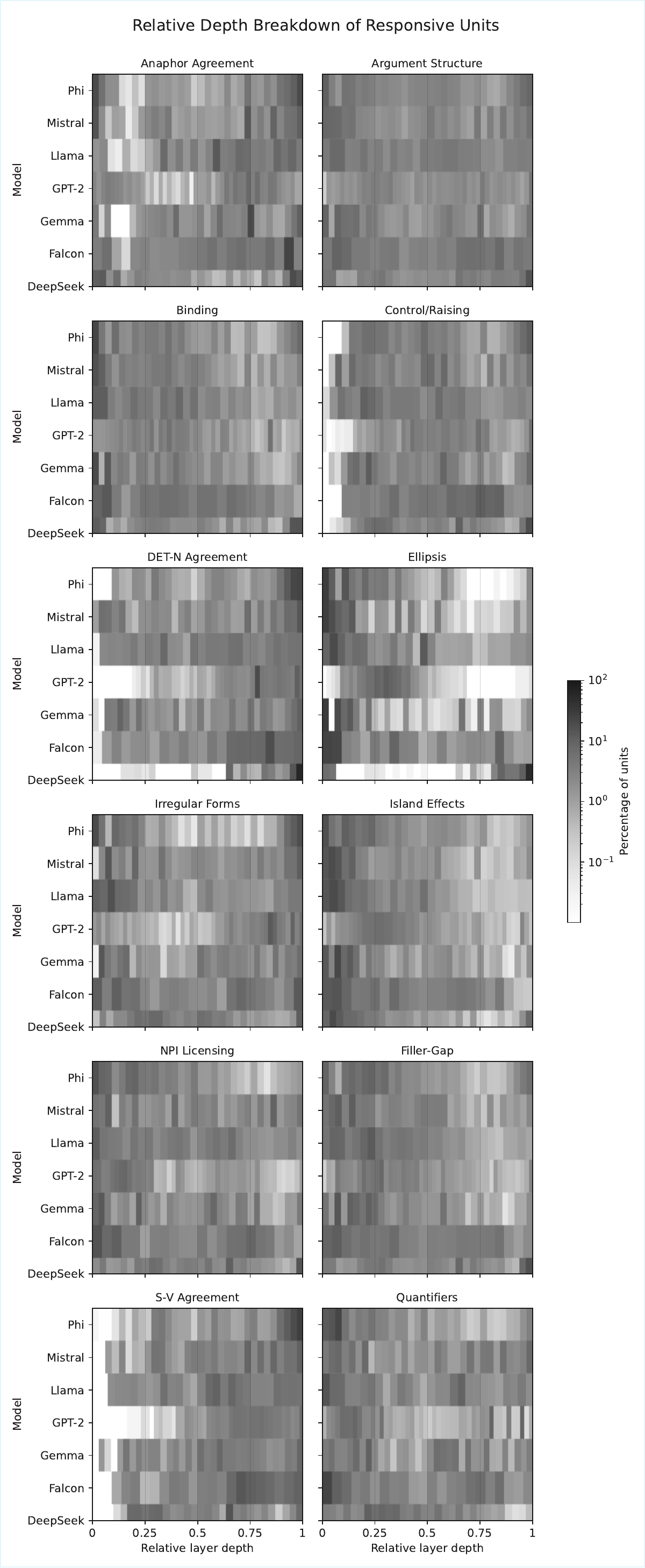}
\caption{\textbf{Breakdown of syntax-responsive units (searched over the attention and MLP modules) by layer.} Each row denotes a model; each bin in model $m$'s row denotes a layer in $m$. The $x$-axis indicates layer depth relative to the whole model (e.g., the 36th layer of a 48-layer model corresponds to $x = 0.75$). The color darkness indicates the percentage of unique units in the given layer out of the total unique units responsive to phenomena from the given category (white = 0\%, black = 100\%).}
\label{fig:layers}
\end{figure}

Besides searching for units over the residual stream, we conducted a more fine-grained search over the outputs of the attention and MLP modules. (The output of the attention module is recorded after out-projection and dropout but before the first residual addition; the output of the MLP module is recorded after down-projection and dropout but before the second residual addition). We analyzed the breakdown of top-1\% units by module type (Figure \ref{fig:mlp-attn}) and by layer (Figure \ref{fig:layers}).

We found that agreement phenomena primarily recruit MLP units, while NPI licensing, binding, and filler-gap phenomena recruit attention units, which suggests that MLPs tend to be involved in local dependencies whereas attention is involved in long-distance dependencies.

For S-V agreement, DET-N agreement, and control/raising, we found the responsive units to be concentrated in the final layers; for ellipsis, island effects, NPI licensing, filler-gap, and quantifiers, in the initial layers; for anaphor agreement, argument structure, binding, and irregular forms, split between the initial and final layers. Contrary to \citet{jawahar-etal-2019-bert}, we did not find many syntactic units in the middle layers.

\section{Five-fold cross-validation}
\label{sec:appendix-5-fold}

\begin{figure}[H]
\includegraphics[width=\columnwidth]{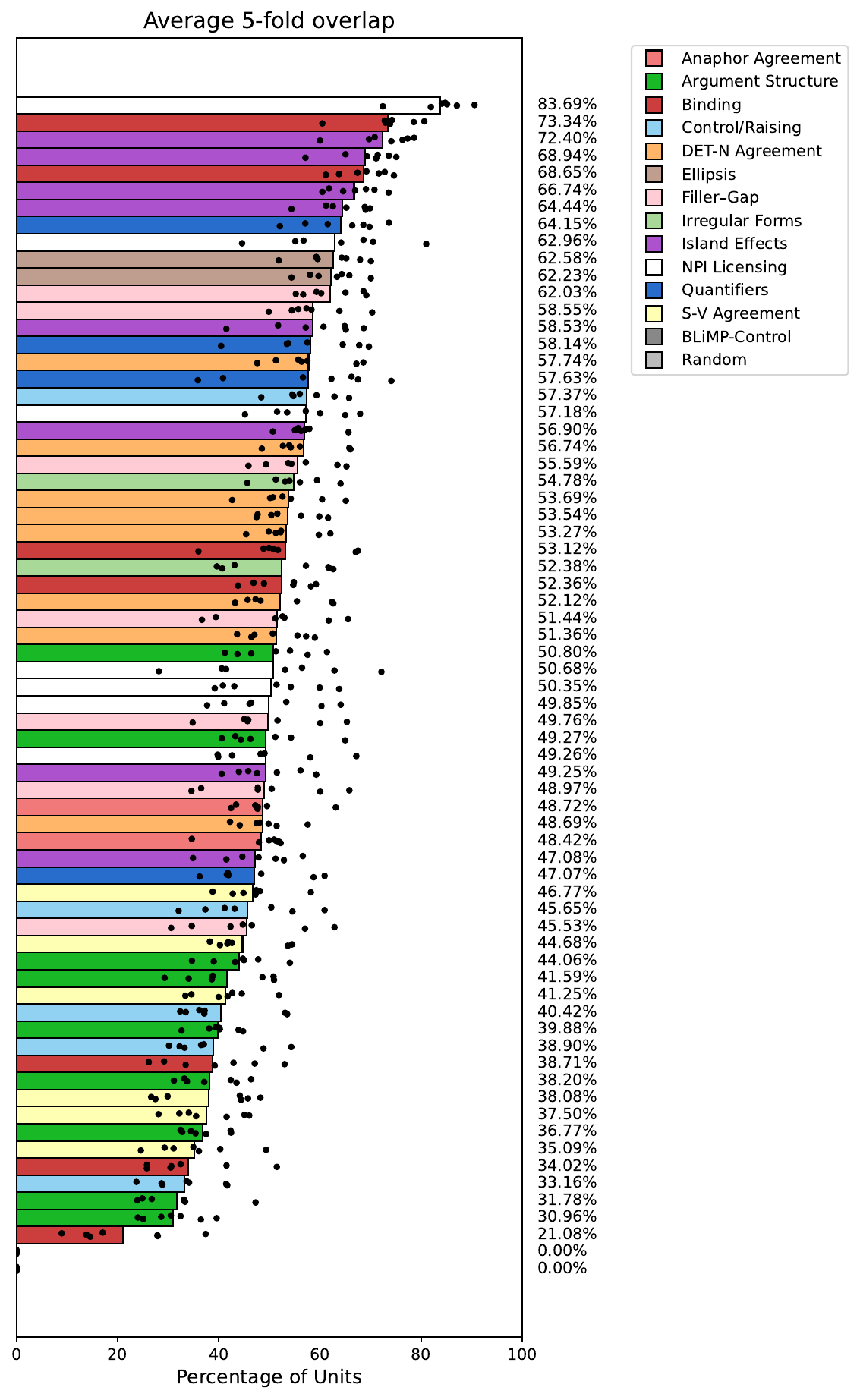}
\caption{Five-fold cross-validation analysis.}
\label{fig:cv-5-fold}
\end{figure}

We performed 5-fold cross-validation by splitting each BLiMP phenomenon into 5 sub-datasets of 200 minimal pairs each, localizing 1\% of units on each sub-dataset, and recording the intersection of the localized unit sets. The results are shown in Figure \ref{fig:cv-5-fold}: syntax-responsive units are still identified reliably, albeit with slightly lower cross-validation scores.

\section{Generalization to other linguistic materials}
\label{sec:appendix-generalization}

Do syntax-responsive LLM units identified from BLiMP generalize to other datasets targeting the same syntactic phenomena, or are they idiosyncratic to BLiMP?
To test this, we applied the same localization procedure to three additional minimal-pair benchmarks: SyntaxGym \cite{gauthier-2020}, the naturalistic subset from \citet{gulordava-2018} (excluding nonce sentences), and the benchmark from \citet{linzen-2016}. The Gulordava and Linzen benchmarks target subject-verb (S-V) agreement only, while SyntaxGym targets multiple categories of phenomena, of which S-V agreement and filler-gap dependencies are shared with BLiMP.

\begin{figure}[H]
\centering
\includegraphics[width=\columnwidth]{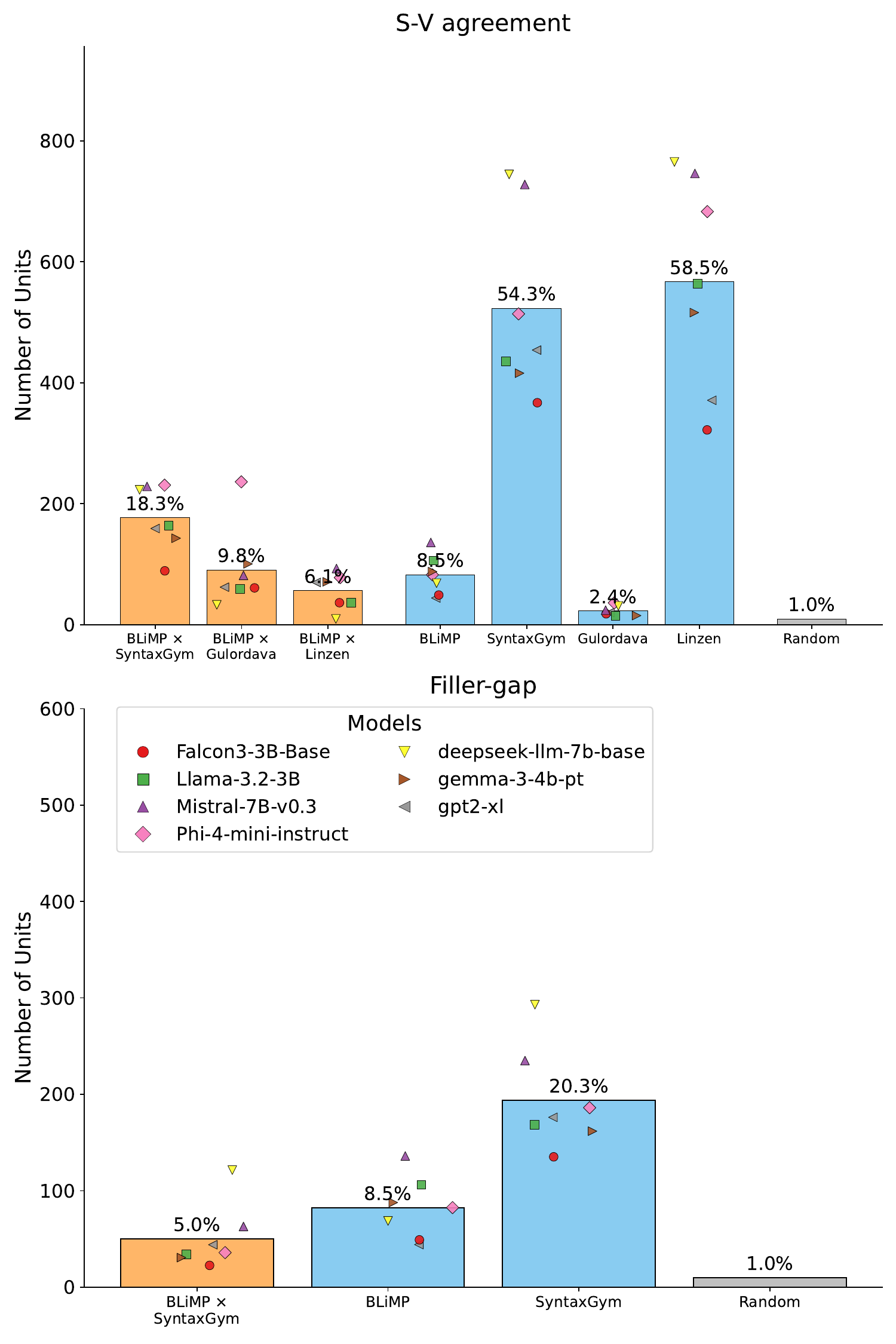}
\caption{\textbf{Generalization of units localized on BLiMP to other materials.} We compare unit overlap across benchmarks, within benchmarks, and the random baseline. Each orange bar shows the average cross-benchmark overlap of the localized unit sets across all possible pairs of phenomena and all models; benchmarks without phenomenon sub-divisions (Gulordava and Linzen) are treated as having one phenomenon. Blue bars denote the average within-benchmark overlap across all possible pairs of phenomena (BLiMP, SyntaxGym) or 2-fold cross-validation overlap (Gulordava, Linzen). Percentages represent the average percentage of overlapping units out of the size of the target 1\% unit set. Markers indicate individual models.}
\label{fig:generalization}
\end{figure}

For S-V agreement, for each new benchmark (SyntaxGym, Gulordava, and Linzen), we computed the overlaps between every possible pair of an S-V agreement phenomenon from BLiMP and an S-V agreement phenomenon from the other benchmark. For filler-gap dependencies, we computed such cross-benchmark overlaps between BLiMP and SyntaxGym. We then compared these overlaps to the within-benchmark consistency: for BLiMP and SyntaxGym, which have multiple phenomena, we computed the overlaps for all possible pairs of two distinct phenomena within the same benchmark and category (S-V agreement/filler-gap), and for Gulordava and Linzen, we performed 2-fold cross-validation.

We found the overlap between the sets of selective units localized on BLiMP and every other benchmark to be significantly higher than random  ($t=2.92$, $p=0.0307$ via a one-sample $t$-test; see Figure \ref{fig:generalization})---even across benchmarks that differ in how the materials were constructed, sentence lengths, and number of sentences. For the Gulordava benchmark, the cross-overlap with BLiMP is even higher than the 2-fold overlap within Gulordava---the low 2-fold consistency is likely due to the small number of minimal pairs (20 per fold; 40 in total).

\section{Mean ablation}
\label{sec:appendix-mean-ablation}

To verify the robustness of the ablation results to different ablation types, we re-ran the ablation experiment from Section~\ref{sec:english_identification_causal} with mean ablation in addition to zero ablation. For each BLiMP phenomenon, we localized top-1\% units using the first 500 sentence pairs, then computed $m$ as the mean over the activations of all units in all layers on the last tokens of the grammatical sentences in this set of pairs (i.e., the underlying premise here is: what happens if we replace a unit’s activation with the value it typically takes for grammatical sentences?). We next set the top-1\% units to $m$ and evaluated accuracy on the held-out 500-pair set.

We observed an average performance drop of 7.88\%, compared to an average drop of 5.82\% obtained when applying mean ablation to random units.

To relate the mean-ablation and zero-ablation results, we computed the Pearson $R$ between the performance drops from the two ablation types. For the majority of models (Llama, Falcon, Phi, DeepSeek, Mistral), we found very high correlations (0.988-0.999). For Gemma, the correlation was somewhat lower (0.766), and for GPT2, it was the lowest (0.541)---yet still highly significant. The models with high correlations have near-zero mean activations (magnitudes on the order of $10^{-4}$--$10^{-5}$), whereas GPT2's mean activations are often significantly different from 0. 

\begin{figure}[H]
\includegraphics[width=\columnwidth]{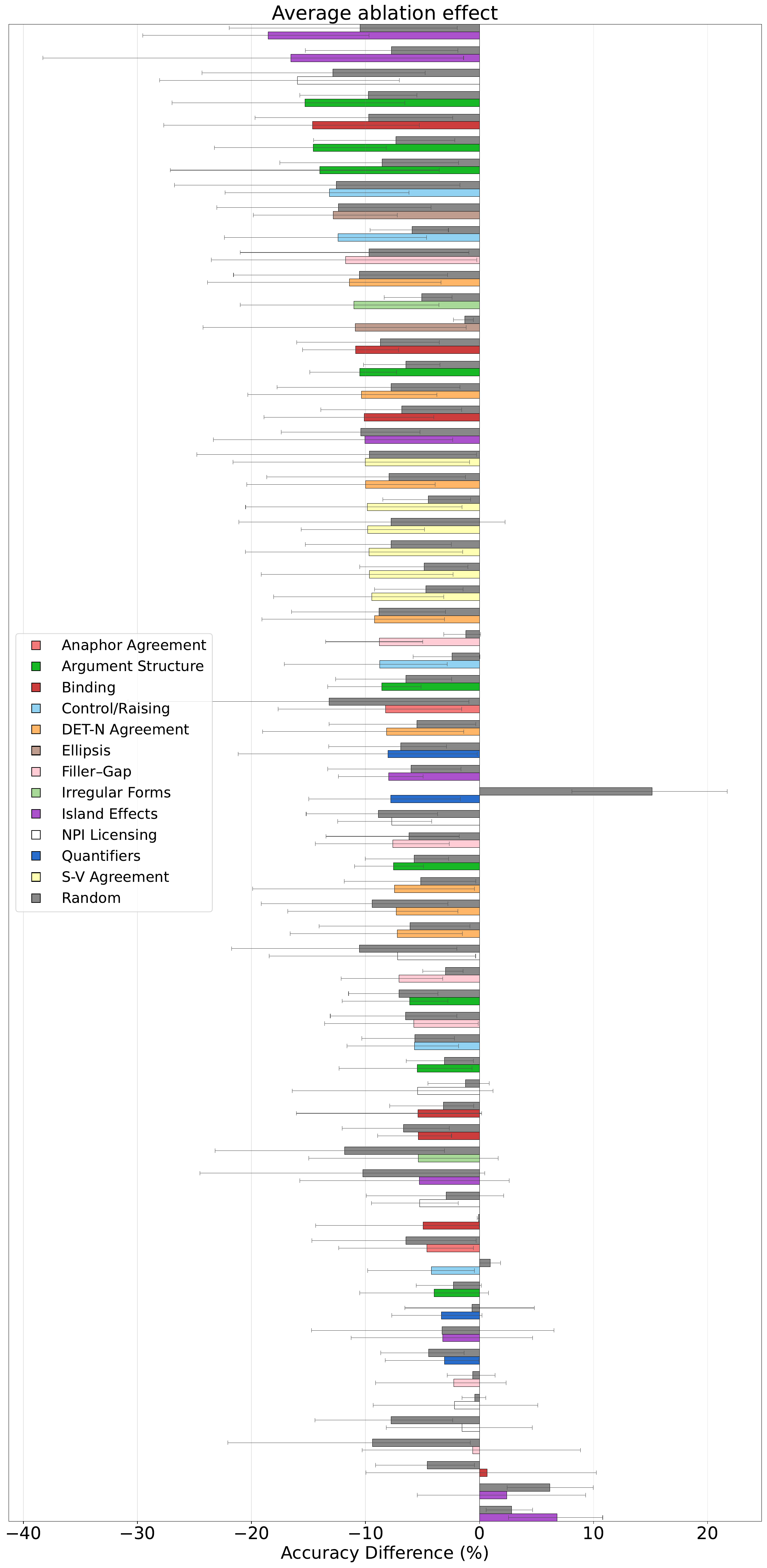}
\caption{Results of the mean ablation experiment.}
\label{fig:appendix-mean-ablation}
\end{figure}

\section{Ablation effect for individual models}
\label{sec:appendix-ablation}

\begin{figure}[H]
\includegraphics[width=0.84\columnwidth]{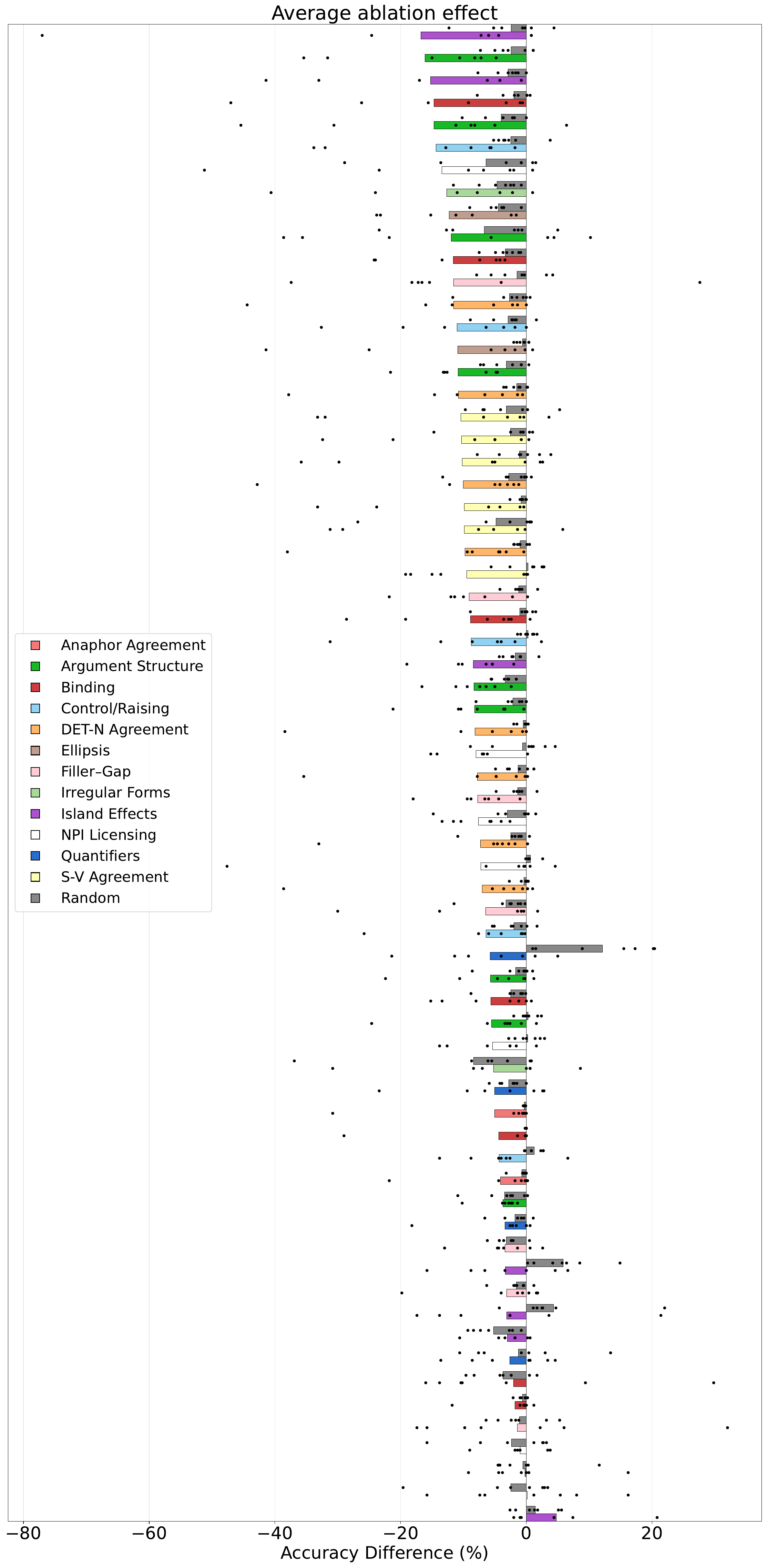}
\caption{Visualization of the results of the ablation experiment from Section \ref{sec:english_identification_causal} with model markers.}
\label{fig:appendix-ablation}
\end{figure}

To demonstrate the variability of ablation-induced performance drops across models, we provide a variant of Figure \ref{fig:ablation} that shows markers for models (Figure \ref{fig:appendix-ablation}).

\section{Correlation between cross-validation consistency and ablation effect}
\label{sec:appendix-scatterplot}

\begin{figure}[H]
\includegraphics[width=0.9\columnwidth]{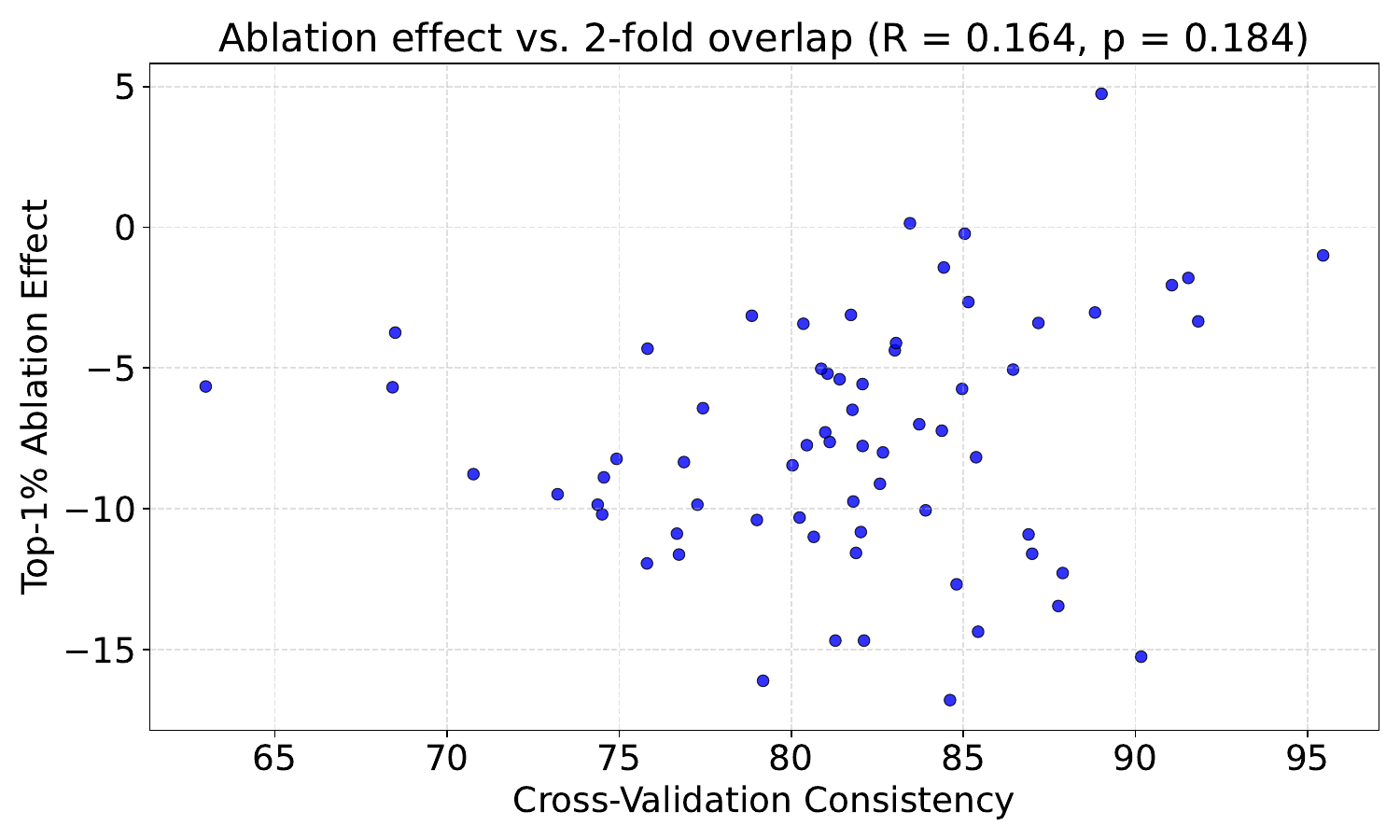}
\caption{\textbf{Scatterplot showing the relationship between cross-validation consistency and ablation effect of the localized unit sets.} Each dot represents one of the 67 BLiMP phenomena; its $x$-coordinate is the average 2-fold cross-validation consistency of the respective top-1\% unit set across the seven models considered, and its $y$-coordinate is the average accuracy difference between the top-unit-ablated and unablated model across the 7 models.}
\label{fig:scatterplot}
\end{figure}

\section{\emph{BLiMP-Lex} control}
\label{appendix-blimp-lex}

For each BLiMP phenomenon, we created two pseudo-phenomena, one with noun substitutions and one with verb substitutions. Each noun pseudo-phenomenon was constructed by copying the grammatical BLiMP sentences from the corresponding phenomenon and deriving their lexical violation counterparts by replacing the first noun with another noun sampled from WordNet. Sampling was matched by frequency ($\pm 1$ unit of Zipf scale) and length ($\pm 3$ characters). An analogous procedure was used to construct the verb pseudo-phenomena. We sub-sampled 15 minimal pairs from each pseudo-phenomenon---to get $\sim$1000 pairs in total---and manually reviewed them to make sure the substitutions did not introduce syntactic violations. We then concatenated and reshuffled the sub-sampled minimal pairs (separately for nouns and verbs). We used the two resulting datasets to localize lexical units and calculate to what extent they overlapped with the syntactic agreement units. Specifically, for each model $m$, for each BLiMP agreement phenomenon $p$, we first computed $s = \frac{n+v}{2}$, where $n$ and $v$ are the percentage overlaps of $p$ with the noun and verb materials respectively. We then averaged the values of $s$ across all agreement phenomena for model $m$, and finally averaged the resulting values across models.

\vspace{300pt}

\section{Cross-validation results for BLiMP, RuBLiMP, and SLING for the Gemma model}
\label{sec:appendix-cv-gemma}

\begin{figure}[b]
\subfigure{
\includegraphics[width=0.66\columnwidth]{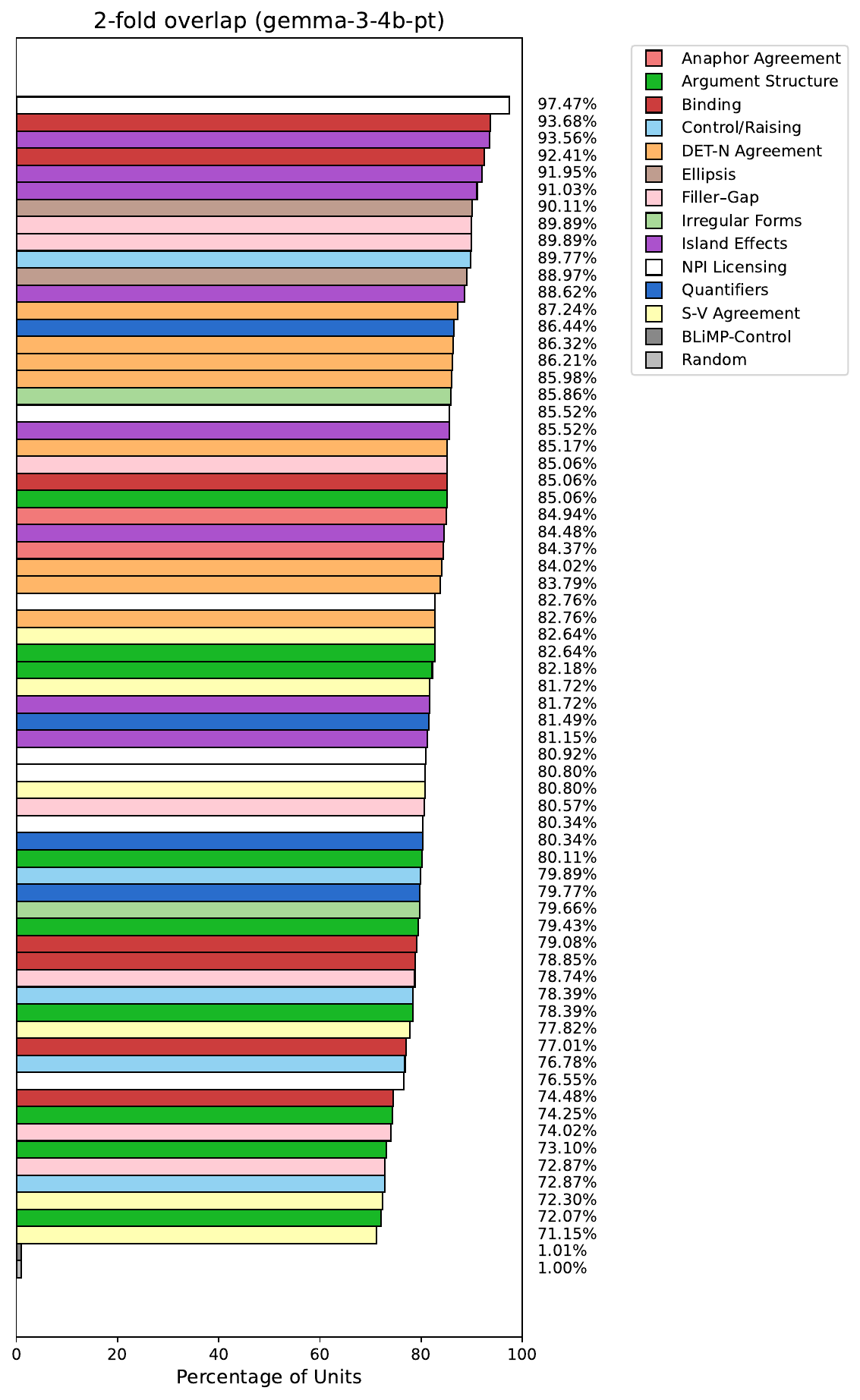}}
\subfigure{
\includegraphics[width=0.77\columnwidth]{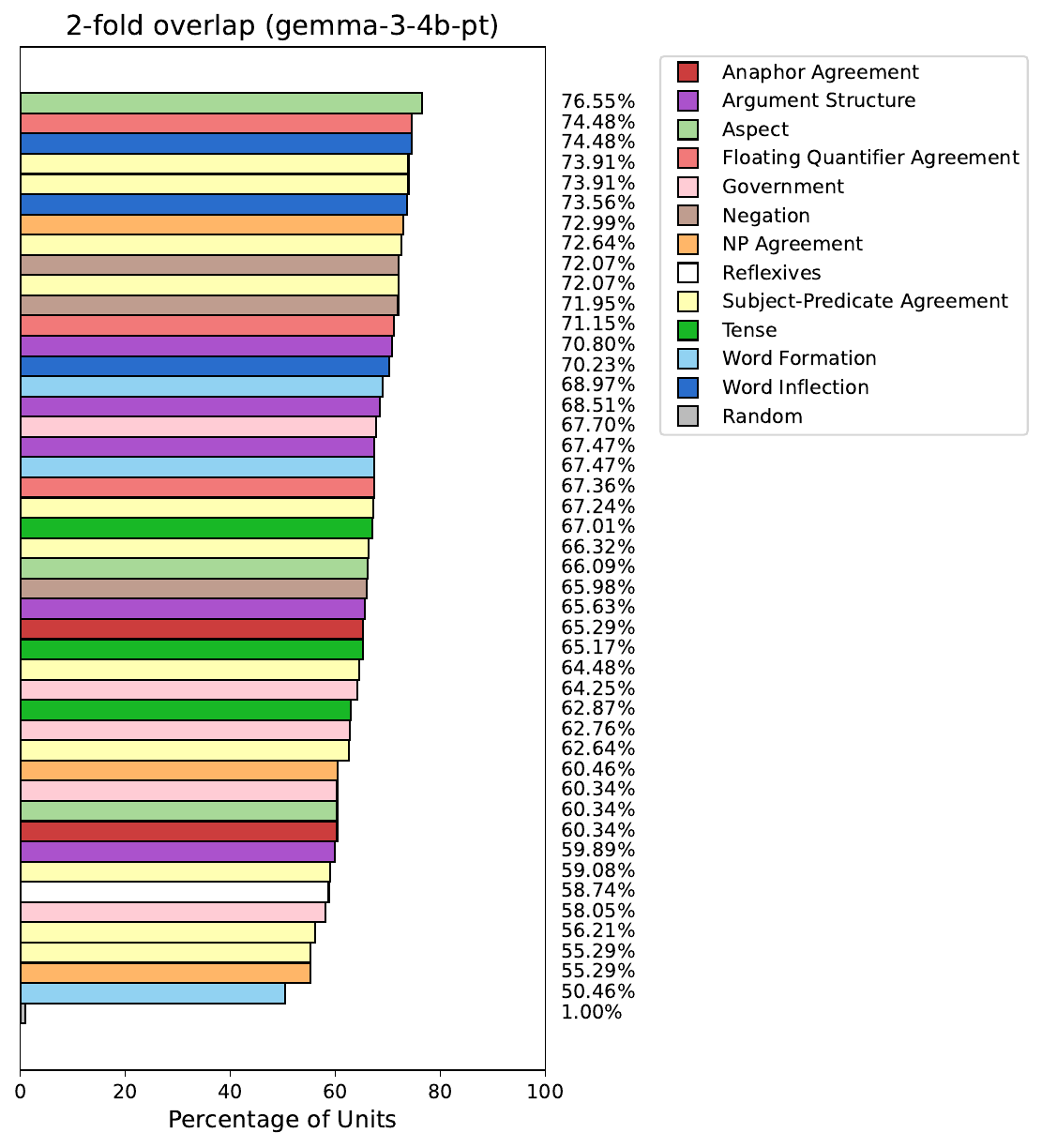}}
\subfigure{
\includegraphics[width=0.76\columnwidth]{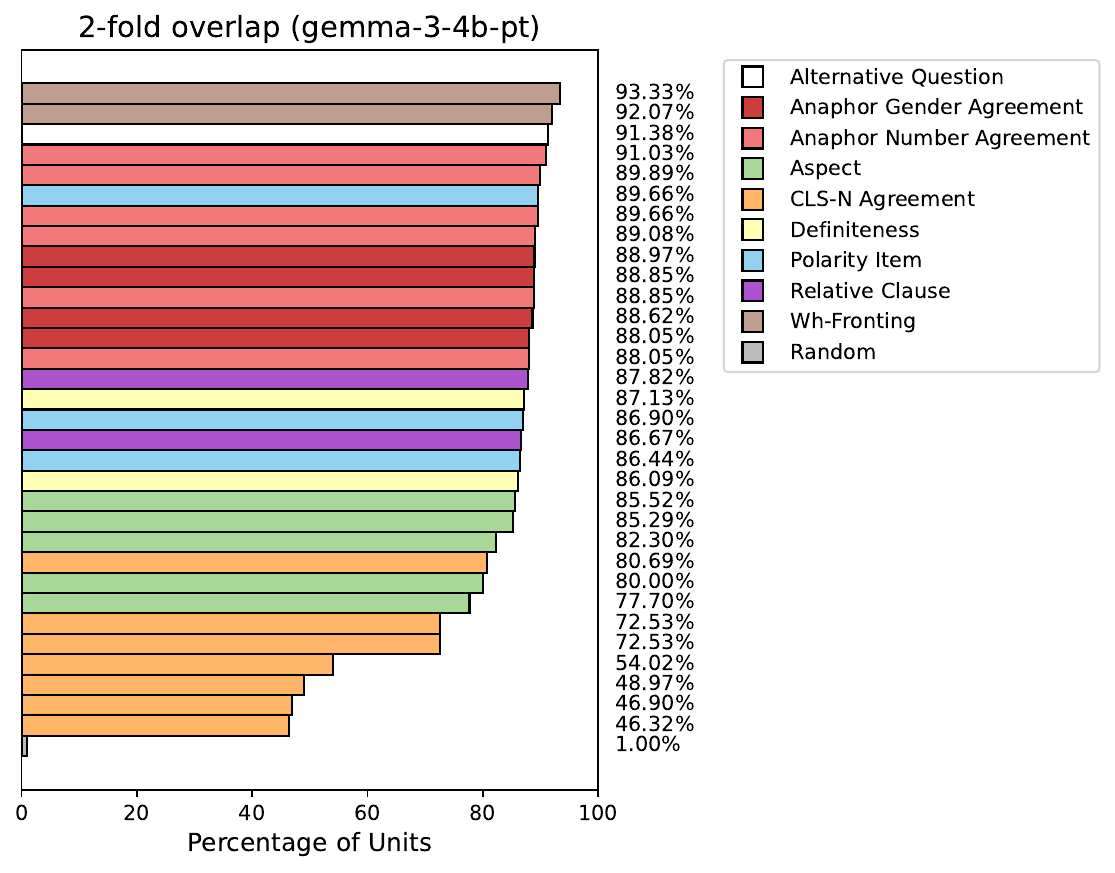}}
\caption{Results of the 2-fold cross-validation analysis for BLiMP (top), RuBLiMP (middle), and SLING (bottom) with the Gemma model. The 2-fold cross-validation consistency is high in these three languages.}
\label{fig:appendix-cv-rublimp-sling}
\end{figure}

\end{document}